\documentclass[10pt,twocolumn,letterpaper]{article}
\usepackage{cvpr}
\usepackage{times}
\usepackage{epsfig}
\usepackage{graphicx}
\usepackage{amsmath}
\usepackage{amssymb}
\usepackage{comment}

\usepackage[table,dvipsnames]{xcolor}
\usepackage{color, colortbl}
\definecolor{Gray}{gray}{0.9}

\usepackage{cite}
\usepackage{xspace}
\usepackage{pifont}
\usepackage[font=normalsize]{subfig}

\usepackage{overpic}
\usepackage{dirtytalk}
\usepackage{booktabs}
\usepackage{multirow}
\usepackage{tabulary}
\usepackage{arydshln}

\newcommand{\deemph}[1]{\textcolor{gray}{#1}}

\newcommand{\cdashlinelr}[1]{%
  \noalign{\vskip\aboverulesep
           \global\let\@dashdrawstore\adl@draw
           \global\let\adl@draw\adl@drawiv}
  \cdashline{#1}
  \noalign{\global\let\adl@draw\@dashdrawstore
           \vskip\belowrulesep}}

\newcolumntype{x}[1]{>{\centering\arraybackslash}p{#1pt}}
\newlength\savewidth

\makeatletter
\renewcommand\paragraph{\@startsection{paragraph}{4}{\z@}%
  {.5em \@plus1ex \@minus.1ex}%
  {-.5em}%
  {\normalfont\normalsize\bfseries}}
\makeatother

\usepackage[pagebackref=true,breaklinks=true,letterpaper=true,colorlinks,bookmarks=false]{hyperref}

\cvprfinalcopy 

\begin{document}

\title{Open Cross-Domain Visual Search}

\author{William Thong \qquad Pascal Mettes \qquad Cees G.M. Snoek \\
University of Amsterdam\\
{\tt\small \{w.e.thong,p.s.m.mettes,cgmsnoek\}@uva.nl}
}

\maketitle

\begin{abstract}
This paper addresses cross-domain visual search, where visual queries retrieve category samples from a different domain. For example, we may want to sketch an airplane and retrieve photographs of airplanes.
Despite considerable progress, the search occurs in a closed setting between two pre-defined domains.
In this paper, we make the step towards an open setting where multiple visual domains are available. This notably translates into a search between any pair of domains, from a combination of domains or within multiple domains.
We introduce a simple -yet effective- approach. We formulate the search as a mapping from every visual domain to a common semantic space, where categories are represented by hyperspherical prototypes.
Open cross-domain visual search is then performed by searching in the common semantic space, regardless of which domains are used as source or target. Domains are combined in the common space to search from or within multiple domains simultaneously. A separate training of every domain-specific mapping function enables an efficient scaling to any number of domains without affecting the search performance.
We empirically illustrate our capability to perform open cross-domain visual search in three different scenarios. Our approach is competitive with respect to existing closed settings, where we obtain state-of-the-art results on several benchmarks for three sketch-based search tasks.
\end{abstract}


\section{Introduction}

This paper aims for visual category search across domains. The task is to retrieve visual examples from a specific category in one domain, given a query from another domain. 
For example, we may want to retrieve \textit{images} of an \say{airplane} from a quickly-drawn \textit{sketch}. Cross-domain visual search has made considerable progress, showing the possibility to retrieve natural images~\cite{eitz2010sketch,sketchy2016} or 3D shapes~\cite{li2013shrec,li2014shrec,li2014comparison} from sketches. Different from existing works, which emphasize retrieval from a single source domain to a single target domain, we open the search beyond two domains. The motivation for a search among many domains is that in practice, categories come in many forms~\cite{peng2018moment,Wilber_2017_ICCV,li2017deeper}. Hence, we may have queries from several source domains, or want to search with any possible combination of source and target domains. For example, we may now want to combine a \textit{sketch} and a \textit{clipart} of an \say{airplane} to retrieve \textit{photograph} samples, or use a \textit{clipart} of an \say{airplane} to retrieve \textit{3D shapes}.
In this paper, we strive for such an open setting: we visually search for categories from any source domain to any target domain, with the ability to search from and within multiple domains simultaneously.

\begin{figure}[t]
\centering
\includegraphics[width=\linewidth]{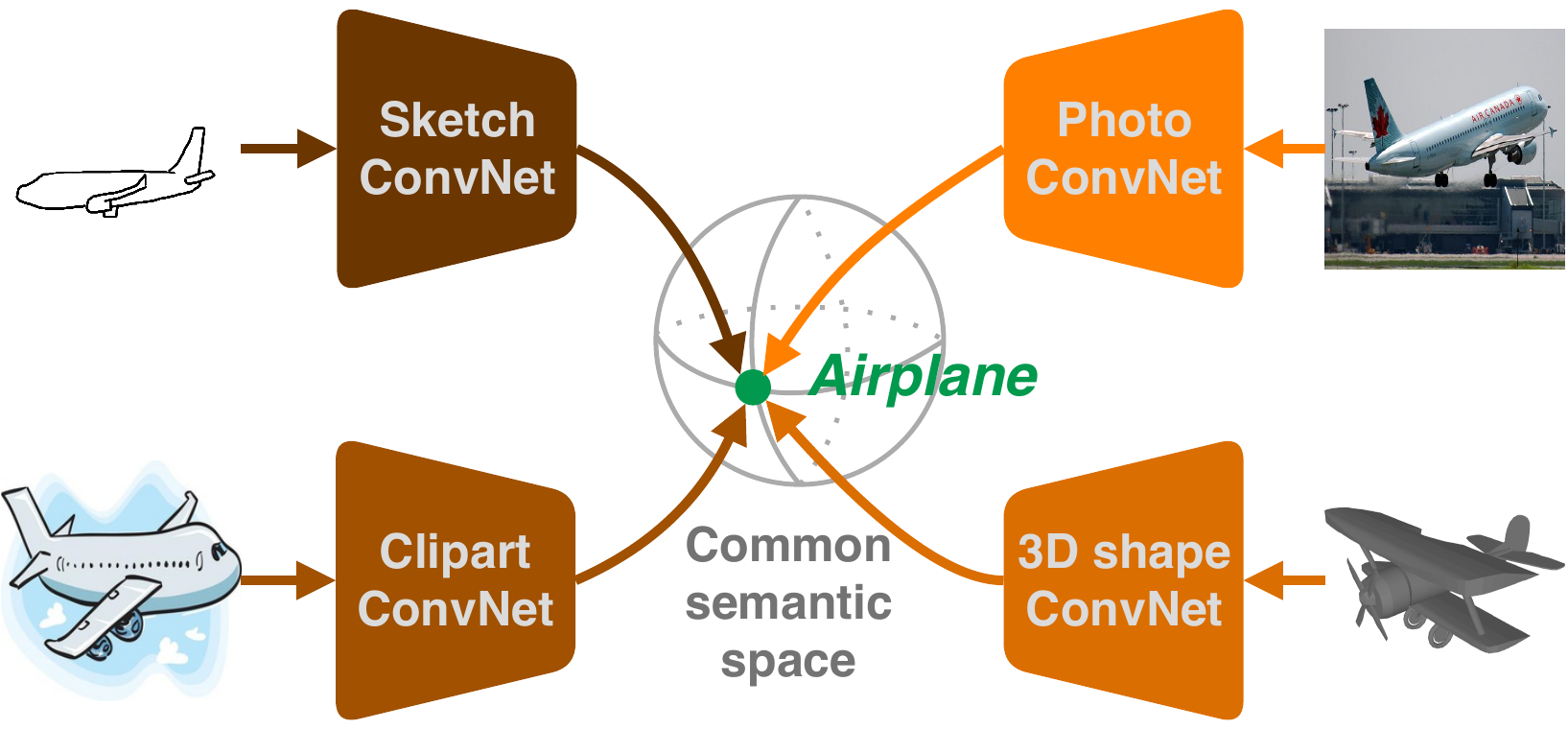}
\caption{\textbf{Open cross-domain visual search.}
We search for categories from any number of source domains to any number of target domains. Mapping examples to a common semantic space enables any possible combinations of domains when searching for categories.\label{fig:one}
}
\end{figure}

Within cross-domain visual search, an important challenge is the gap between source and target domains~\cite{shen2018zero,kiran2018zero,dey2019doodle,dutta2019semantically,xie2017learning,chen2019deep}. Given the inherent difference in representations, reducing the domain gap is an intuitive solution. Both Shen~\etal~\cite{shen2018zero} and Yelamarthi~\etal~\cite{kiran2018zero} have highlighted the importance of domain adaptation losses for cross-domain search, especially when searching for unseen categories. Yet, relying on domain adaptation methods makes the search unsuited for an open setting by design, due to the requirement of pair-wise domain training. As a consequence, opening the search to many domains creates new challenges as (\textit{i}) all domains should to be mapped to a unique embedding space, and (\textit{ii}) new domains should be able to be added continuously in an efficient fashion. We address the challenges of open cross-domain visual search.

Inspired by recent works on prototype-based embedding spaces~\cite{movshovitz2017no,wen2016discriminative,snell2017prototypical}, we introduce prototype learners for cross-domain visual search in an open setting. Prototype learning has shown to simplify model training and improve performance for image retrieval~\cite{movshovitz2017no,wen2016discriminative} and classification~\cite{snell2017prototypical} problems in a low-shot setting. In this work, we leverage prototype learners to perform visual search across multiple domains simultaneously.
We define prototypes to unite all domains. Inputs from every domain are mapped to a common semantic space, where every learner is domain-specific and is trained separately. During training, the semantic space is defined by categorical prototypes, corresponding to word embeddings of category names. Learning then consists of regressing inputs to their corresponding categorical prototype in this common semantic space, as illustrated in Figure~\ref{fig:one}. Query representations for search are further refined with neighbours from other domains through a spherical linear interpolation operation. Once trained, the proposed formulation allows us to search among any pair of domains. Since all domains are now aligned in the common semantic space, this enables a search from multiple source domains or in multiple target domains. Lastly, new domains can be added on-the-fly, without retraining previous models.

Empirically, we first demonstrate the ability to perform open cross-domain visual search, highlighting new applications and search possibilities, \ie~(\textit{i}) a search between any pair of source and target domains without hassle;
(\textit{ii}) a search from multiple source domains;
and (\textit{iii}) a search in multiple target domains. Second, while designed for the open cross-domain setting, our approach also works in the conventional closed settings, allowing for comparisons to current approaches. We compare to sketch-based image and 3D shape retrieval, usually considered separately in the literature. We show the versatility of our approach to handle them. Across three well-established tasks totalling seven benchmarks, we obtain state-of-the-art results, highlighting the effectiveness of focusing solely on the semantic space for cross-domain search.

\paragraph{Contributions} Our main contribution is the introduction of open cross-domain visual search. We open the search to many domains, with the ability to retrieve categories from and among any number of domains. To achieve this, we introduce a simple prototype learner for each domain to learn a common semantic space efficiently. Empirically, solely relying on semantic prototypes turns into an effective solution for cross-domain visual search in both newly proposed open settings and existing closed settings. All code and setups are released to foster further research in open cross-domain visual search\footnote{Source code is available at \href{https://github.com/twuilliam/open-search}{https://github.com/twuilliam/open-search}}.

\section{Related Work}

We first cover related work in cross-domain search, where a large body of works focuses on retrieving natural images or 3D shapes from sketches. We then review relevant work addressing multiple domains and on how to learn semantic spaces with prototype learners.

\paragraph{Cross-domain image search} Sketch-based image retrieval has been a topic of vision community interest for a long time~\cite{kato1992database,jacobs1995fast}. The seminal work of Eitz~\etal~\cite{eitz2010sketch} established the first benchmark for its evaluation, which led to the construction of common descriptors for sketches and images, such as bag-of-features~\cite{eitz2010sketch}, bag-of-regions~\cite{hu2011bag}, histogram of oriented gradients~\cite{hu2013performance}, or specialized descriptors for edges~\cite{saavedra2014sketch}. With the resurgence of convolutional networks, the dominant approach has shifted towards the learning of a joint semantic space of sketches and images. Qi~\etal\cite{qi2016sketch} learn a joint embedding with a Siamese network while Bui~\etal\cite{bui2017compact} rely on a triplet network. Bui~\etal\cite{bui2018sketching} add a classification head with a multi-stage training to make features even more discriminative. In all these works, the semantic spaces model categories implicitly, as they rely on sample-based methods such as the Siamese~\cite{hadsell2006dimensionality,chopra2005learning} or triplet~\cite{schroff2015facenet,weinberger2009distance} losses to learn cross-domain visual similarities. In this paper, we explicitly define semantic representations for every category in the embedding space. This removes the need for sampling and mining of cross-domain pairs, resulting in a much simpler training procedure.

\begin{figure*}[t]
\vspace{-1em}
\centering
\subfloat[][{\small\centering\textit{one} source to \textit{one} target}\label{fig:two-d}]{
\begin{overpic}[width=0.22\linewidth, trim={0px -43px 0px 0px}, clip]{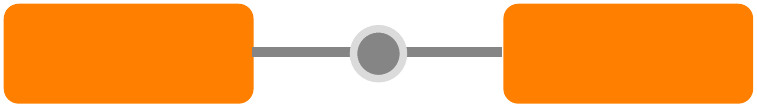}
\put(15,24){$s$}
\put(82,24){$t$}
\end{overpic}
}\hfill
\subfloat[][ {\small\centering\textit{any} source to \textit{any} target}\label{fig:two-a}]{
\begin{overpic}[width=0.22\linewidth]{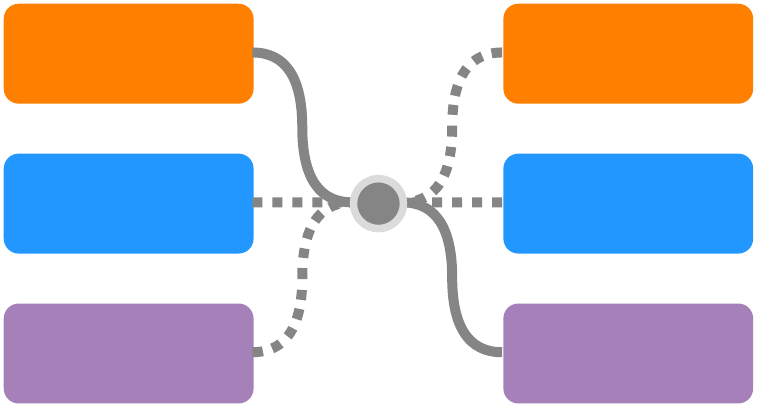}
\put(13,44){$s_1$}
\put(80,44){$t_1$}
\put(13,25){$s_2$}
\put(80,25){$t_2$}
\put(78,15){$\cdots$}
\put(12,15){$\cdots$}
\put(13,5){$s_K$}
\put(80,5){$t_K$}
\end{overpic}
}\hfill
\subfloat[][ {\small\centering\textit{many} sources to \textit{any} target}\label{fig:two-b}]{
\begin{overpic}[width=0.22\linewidth]{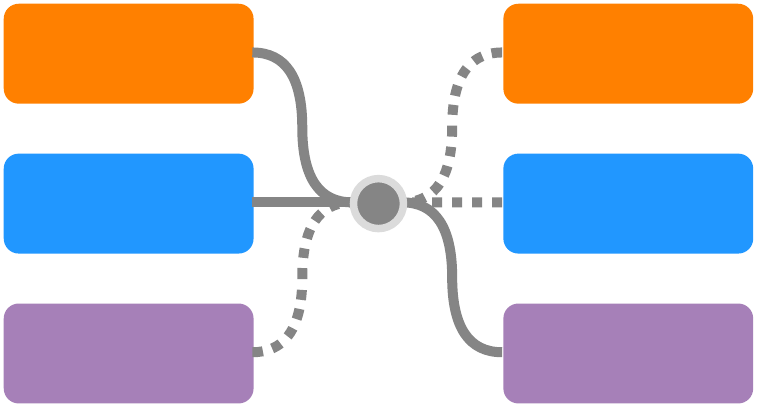}
\put(13,44){$s_1$}
\put(80,44){$t_1$}
\put(13,25){$s_2$}
\put(80,25){$t_2$}
\put(78,15){$\cdots$}
\put(12,15){$\cdots$}
\put(13,5){$s_K$}
\put(80,5){$t_K$}
\end{overpic}
}\hfill
\subfloat[][{\small\centering\textit{any} source to \textit{many} targets}\label{fig:two-c}]{
\begin{overpic}[width=0.22\linewidth]{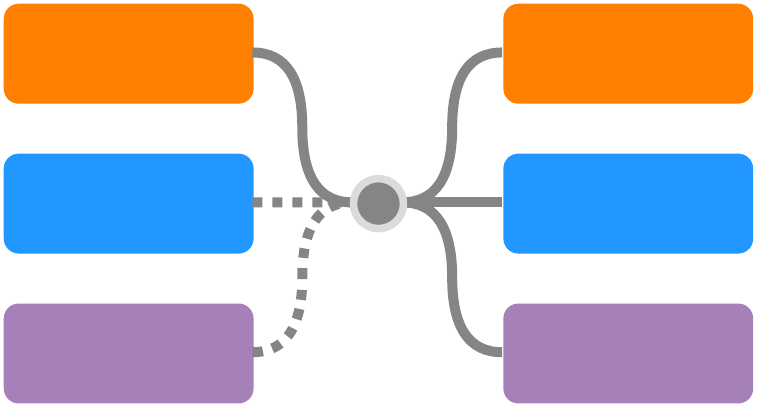}
\put(13,44){$s_1$}
\put(80,44){$t_1$}
\put(13,25){$s_2$}
\put(80,25){$t_2$}
\put(78,15){$\cdots$}
\put(12,15){$\cdots$}
\put(13,5){$s_K$}
\put(80,5){$t_K$}
\end{overpic}
}
\caption{\textbf{Open cross-domain visual search configurations.}
Cross-domain image search focuses on mapping (a) from one fixed source to one fixed target domain. In this paper, we consider an open domain setting with $K$ available domains. We search (b) from any source to any target domain, (c) from multiple source domains to any target domain, and (d) from any source domain to multiple target domains.
}
\label{fig:two}
\vspace{-1em}
\end{figure*}
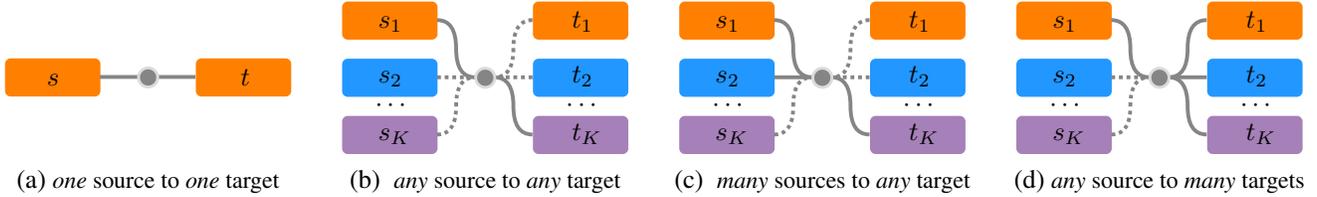

Sketch-based image retrieval is also considered as a zero-shot learning problem ~\cite{shen2018zero,kiran2018zero}. In this context, a common approach is to bridge the domain gap between sketches and images.
Shen~\etal\cite{shen2018zero} fuse sketch and image representations with a Kronecker product, while Yelamarthi~\etal\cite{kiran2018zero} introduce domain confusion with generative models to produce domain-agnostic features.
Dey~\etal\cite{dey2019doodle} combine gradient reversal layers with metric learning losses to extract the mutual information from both domains.
Duttan and Akata\cite{dutta2019semantically} tie the semantic space with visual features from both domains by learning to generate them while Dutta and Biswas~\cite{dutta2019style} prefer to separate them explicitly.
Alternatively, Liu~\etal\cite{liu2019semantic} preserve the knowledge from a pre-trained model to avoid features to drift away during training.
Hu~\etal\cite{hu2018sketch} have also explored how to synthesize classifiers derived from sketches for few-shot image classification.
By focusing on domain adaptation, current approaches are optimized to map from a single specific source domain to a single specific target domain. Instead, we consider cross-modal image search from any number of source domains to any number of target domains.

\paragraph{Cross-domain 3D shape search}
Searching for 3D shapes from a sketch has been accelerated by the SHREC challenges~\cite{li2013shrec,li2014shrec,li2014comparison}.
A common approach is to transform the 3D shape search into an image search problem by projecting the unaligned 3D shape into multiple 2D views~\cite{su15mvcnn}. In this regard, the main methodological approach is to learn a joint embedding space of sketches and 2D view renderings of the unaligned 3D shapes.
Wang~\etal\cite{wang2015sketch} map both sketches and shapes in a similar feature space with a Siamese network, while Tasse and Dogson\cite{Tasse2016shape} learn to regress to a semantic space with a ranking loss.
Dai~\etal\cite{dai2017deep} correlate both sketch and 3D shape representations to bridge the domain gap.
Xie~\etal\cite{xie2017learning} employ the Wasserstein distance to create a barycentric representation of shapes.
Qi~\etal\cite{Qi2018SemanticEF} apply loss functions on the probabilistic label space rather than the feature space. 
Chen~\etal\cite{chen2019deep} propose an advanced sampling of 2D views for the unaligned shapes.
Learning cross-domain visual similarities with Siamese or triplet losses typically requires a multi-stage training or negative sampling schemes. A prototype learner removes this requirement, and enables the addition of new domains without the need for retraining existing models.

\paragraph{Searching beyond two domains}
Using multiple domains has been investigated in unsupervised domain adaptation~\cite{peng2017visda,csurka2017domain} and unsupervised domain generalization~\cite{blanchard2011domgen}, where the task is to classify unlabeled target samples by learning a classifier on labeled source samples.
As such, Peng~\etal\cite{peng2018moment} illustrate how challenging classification becomes when multiple domains are considered.
A new challenge then arises as classifiers have to be designed to benefit from the inherent gap among multiple domains~\cite{xu2018deep,peng2018moment,Zhuo_2019_CVPR,dou2019domain,carlucci2019domain}.
In this paper, we focus on a different multi-domain task: we consider cross-domain retrieval where category labels are present for both source and target domains, and where the main challenge is to learn a common embedding space for all domains.

\paragraph{Prototype learners} Learning metric spaces with prototypes for image retrieval~\cite{NIPS2016_6200,wen2016discriminative,movshovitz2017no,zhai2018making,liu2017sphereface,wang2018cosface,deng2019arcface,snell2017prototypical} and classification~\cite{mensink2013distance,snell2017prototypical,chintala2017scale,mettes2019hyperspherical} provides a simpler alternative to common contrastive~\cite{hadsell2006dimensionality,chopra2005learning} or triplet~\cite{schroff2015facenet,weinberger2009distance} loss functions.
One line of work learns to regress to moving prototypical representations. Depending on the task, such prototypes can correspond to center~\cite{wen2016discriminative}, proxy~\cite{movshovitz2017no,zhai2018making}, or support~\cite{snell2017prototypical,ren2018meta} representations. While the distance measure usually relies on a cosine or Euclidean distance, a margin has also been introduced in the distance measure~\cite{liu2017sphereface,wang2018cosface,deng2019arcface}.
Another line of work regresses to fixed prototypical representations to avoid the simultaneous learning of prototypes and model parameters. Examples of fixed representations include class means~\cite{mensink2013distance}, one-hot representations~\cite{chintala2017scale}, or separated representations~\cite{mettes2019hyperspherical}.
We build on the latter approach for open cross-domain visual search. We formulate semantic prototypes to align examples from many domains simultaneously. Categories are represented by fixed semantic prototypes in the embedding space.
We then define a prototype learner for every domain to map visual inputs to the common space where open cross-domain search occurs.


\section{Method}

\subsection{Problem formulation}
Figure~\ref{fig:two} illustrates the search scenarios for open cross-domain search.
While the \textit{closed} cross-domain setting focuses on one pre-defined source $s$ and one pre-defined target $t$, the \textit{open} cross-domain setting searches for categories from any source domain $s_k$ to any target domain $t_k$. As multiple domains now become available, this opens the door for combining multiple domains at both source and target positions.
Thus, the main difference between the \textit{closed} setting and the \textit{open} setting lies in the ability to leverage multiple domains for categorical cross-domain visual search.

Formally, let $\mathcal{D}$ denote the set of all domains to be considered. Rather than making an explicit split of a dataset into source and target, we consider a large combined visual collection
$\mathcal{T} = \{(\mathbf{x}^d_n,y_n)\}_{n=1}^{N}$, where $\mathbf{x}^d_n \in \mathcal{I}_d$ denotes an input example from a visual domain $d \in \mathcal{D}$ of category $y_n \in \mathcal{Y}$. In other words, $\mathcal{Y}$ is common and shared among all domains $\mathcal{D}$ but is depicted differently from domain $d_i$ to domain $d_j$, with $i \neq j$.

Categorical search consists in using a sample query $\mathbf{x}^{d_i}$ from domain $d_i$ to retrieve samples of the same category $y$ in the gallery of domain $d_j$. If $i \neq j$, this corresponds to a cross-domain categorical search as the search occurs across two different domains. A \textit{closed} setting only considers $|\mathcal{D}|=2$, \ie~with a pre-defined source domain and a  pre-defined target domain. We define the \textit{open} setting as comprising $|\mathcal{D}|>2$. This stimulates novel search configurations. For example, we may want to combine two queries $(\mathbf{x}^{d_i}, \mathbf{x}^{d_j})$ of two different domains $i\neq j$ to search in the gallery of a third domain $k$. Conversely, given a sample query $\mathbf{x}^{d_i}$, we can search in the combined gallery of multiple domains.

\subsection{Proposed approach}
We pose open domain visual search as projecting any number of heterogeneous domains to prototypes on a common and shared hyperspherical semantic space. First, we outline how to represent categories in the semantic embedding space. Second, we propose a mapping function for every domain to the common semantic embedding space. Third, we outline how open cross-domain search occurs.

\paragraph{Categorical prototypes}
We leverage the concept of prototypes to represent categories in a common semantic space. Every category is represented by a unique real-valued vector, corresponding to a categorical prototype. Hence, the objective is to align examples, coming from different domains but with the same category label, to the same categorical prototype in the semantic space.
For every category $y \in \mathcal{Y}$, we denote its prototype on the semantic space as $\phi(y) \in \mathbb{S}^{D-1}$ for a $D$-dimensional hypersphere.
Relying on semantic relations enables to search for unseen classes using models trained on seen categories~\cite{frome2013devise,palatucci2009zero}. In this work, we opt for word embeddings, \eg, word2vec~\cite{mikolov2013efficient} or GloVe~\cite{pennington2014glove}, to represent categories, as these embeddings adhere to the semantic relation property.

\paragraph{Mapping domains to categories}
For every domain $d \in \mathcal{D}$, we learn a separate mapping function $f_d(\cdot) \in \mathbb{S}^{D-1}$ to the common and shared semantic space. Separate mapping functions are not only easy to train, they also enable us to incorporate new domains over time. Indeed, we only have to train the mapping of the new incoming domain without retraining previous mapping functions of existing domains. The mapping function is formulated as a convolutional network followed by an $\ell_2$-normalization on the $D$-dimensional network outputs.

We propose the following function to map an example $\mathbf{x}^d$ of domain $d$ to its categorical prototype $\phi(y)$ in the common semantic space:
\begin{equation}
\label{eq:py}
p(y|\mathbf{x}^d,d) = \frac{\exp\Big(-s \cdot c\big(f_d(\mathbf{x}^d), \phi(y)\big)\Big)}{\sum_{y' \in \mathcal{Y}}\exp\Big(-s \cdot c\big(f_d(\mathbf{x}^d), \phi(y')\big)\Big)},
\end{equation}
where $s \in \mathbb{R}_{>0}$ denotes a scaling factor, inversely equivalent to the temperature~\cite{hinton2014distilling}. Intuitively, the scaling controls how samples are spread around categorical prototypes.
$c(\cdot,\cdot)$ is defined as the cosine distance:
\begin{equation}
c(f_d(\mathbf{x}^d), \phi(y)) = 1 - <\!f_d(\mathbf{x}^d), \phi(y)\!>,
\end{equation}
where $<\!\cdot, \cdot\!>$ is the dot product. As both $f_d(\mathbf{x})$ and $\phi(y)$ lie on the hypersphere $\mathbb{S}^{D-1}$, they have a unit norm. Finally, learning every mapping function $f_d$ is done by minimizing the cross-entropy loss over the training set:
\begin{equation}
\label{eq:loss}
\mathcal{L} = - \frac{1}{N}\sum_{n=1}^{N} \log p(y_n|\mathbf{x}^d_n, d).
\end{equation}
In our approach, the representations of the categorical prototypes remain unaltered. Hence, we only take the partial derivative with respect to the mapping function parameters. When training the mapping function $f_d$ for domain $d$, only examples $\mathbf{x}^d$ of domain $d$ are used as inputs.

\paragraph{Searching across open domains}
In the search evaluation phase, similarity between source and target samples is measured with the cosine distance in the shared semantic space. Given one or more queries from different source domains, we first project all queries to the shared semantic space and average their positions into a single vector. Then, we compute the distance to all target examples to rank them with respect to the source query.
As all domains map to the same common semantic space, domains can straightforwardly be combined either to search with queries from multiple domains or to search within a gallery of multiple domains.

\begin{figure}[t]
\vspace{-1em}
\centering
\hfill
\subfloat[ Ideal.]{
\includegraphics[width=0.3\linewidth]{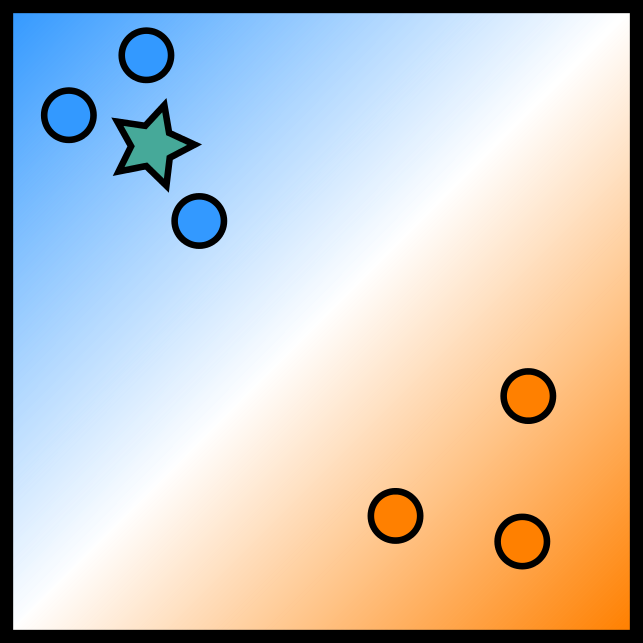}\label{fig:ref:a}}\hfill
\subfloat[ Real.]{
\includegraphics[width=0.3\linewidth]{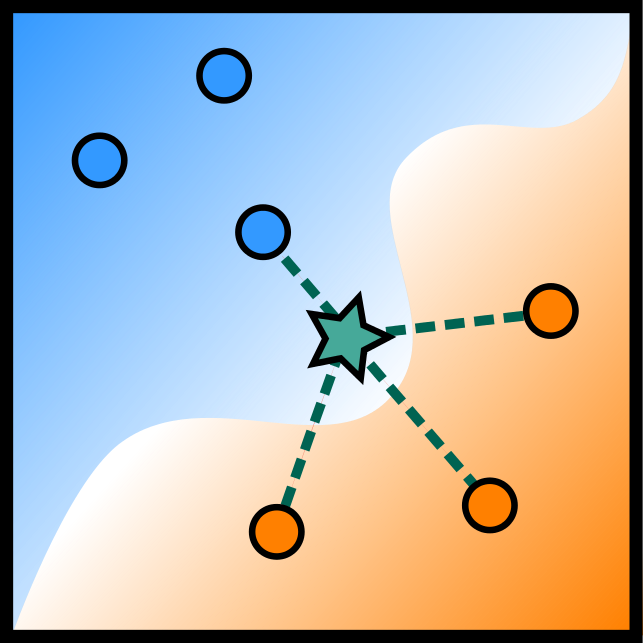}\label{fig:ref:b}}\hfill
\subfloat[ Refined.]{
\includegraphics[width=0.3\linewidth]{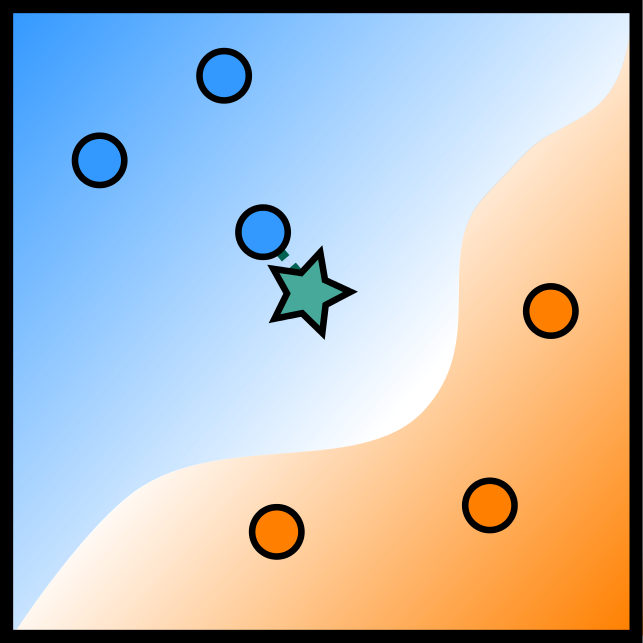}\label{fig:ref:c}}\hfill\null
\caption{\textbf{Cross-domain query refinement}. (a) Ideally, the neighborhood of the query (star) is only close to examples from the same category. (b) In reality, variability causes noise in the semantic space. Hence, the query might also be close to samples from other categories. (c) We tackle this variability by refining the query representation.}
\label{fig:refinement}
\vspace{-1em}
\end{figure}

\begin{figure*}[t]\centering
\vspace{-1em}
\hfill
\subfloat[][\centering \textit{Zero}-shot evaluation (on 45 unseen classes).]{
\includegraphics[height=16em]{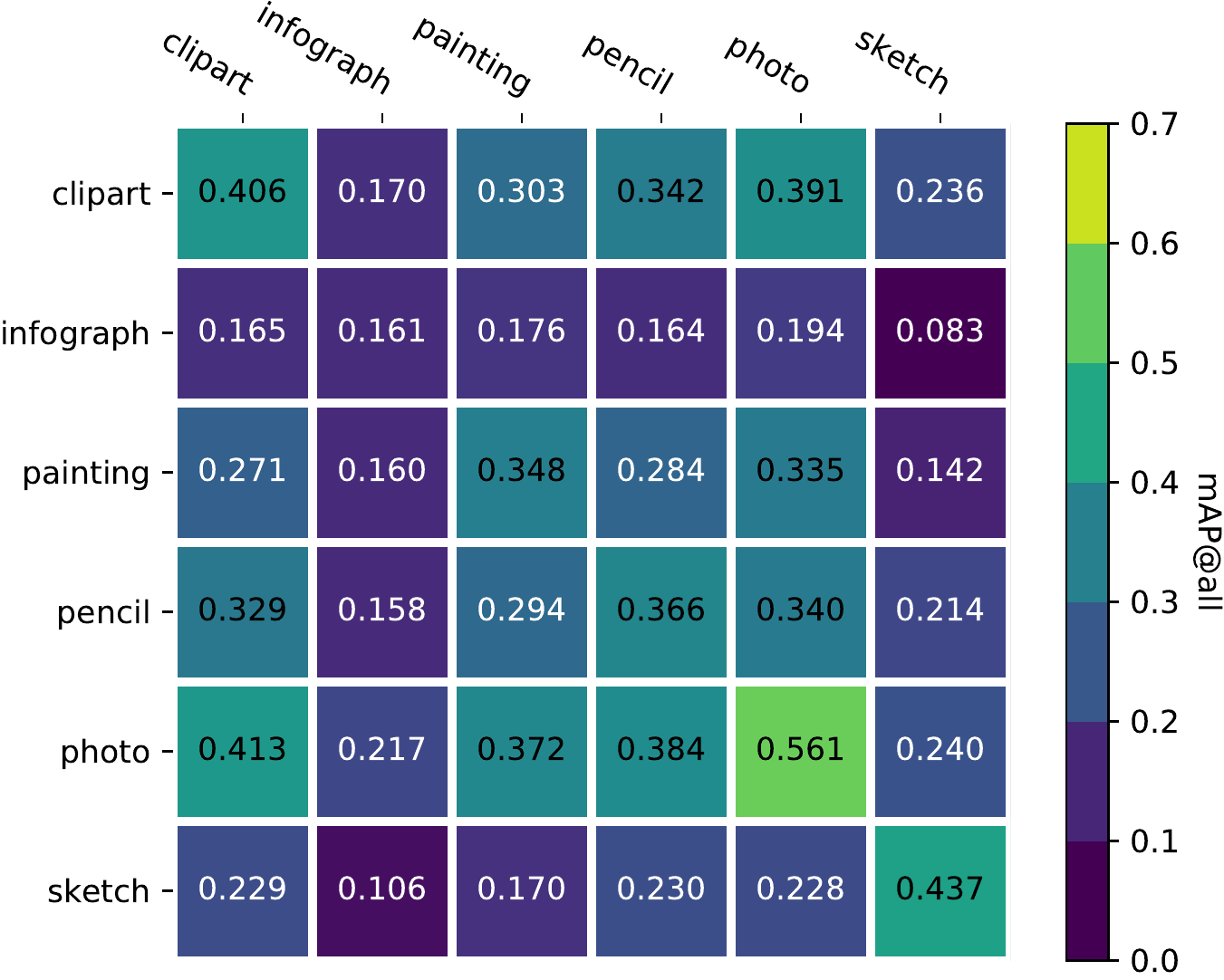}\label{fig:anya}}\hfill
\subfloat[][\centering \textit{Many}-shot evaluation (on all 345 classes).]{
\includegraphics[height=16em]{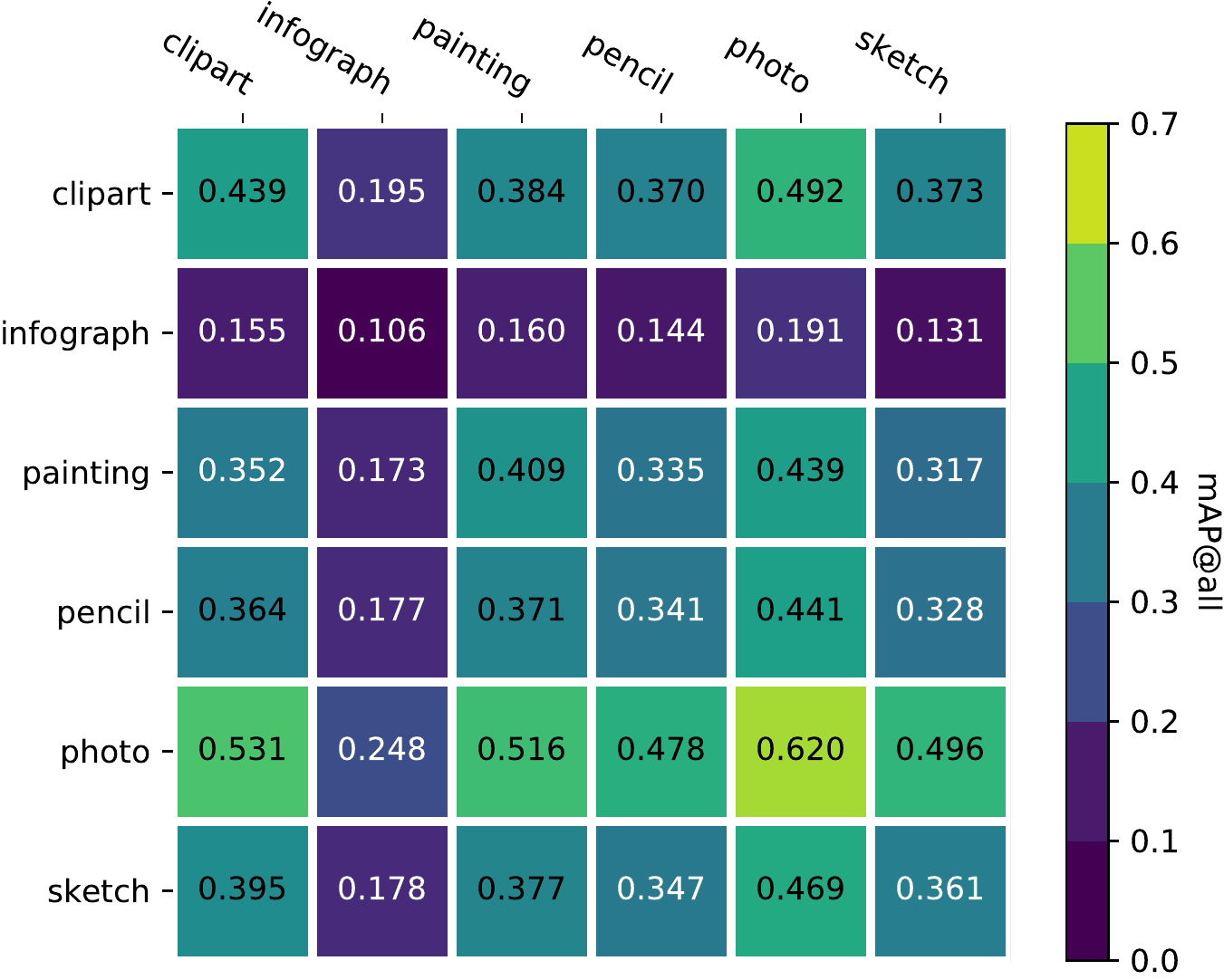}\label{fig:anyb}}\hfill\null
\caption{\textbf{Demonstration 1} for visual search from any source (columns) to any target (rows) domain in mAP@all. Our approach can perform 36 cross-domain searches for both (a) \textit{zero}-shot evaluation, and (b) \textit{many}-shot evaluation, without any modifications as we bypass the need to align domains.
}
\label{fig:any}
\vspace{-1em}
\end{figure*}

\subsection{Refining queries across domains}
With our approach, a source query is close to target examples from the same category, regardless of the domains of the query and target examples. In practice, inherent variability in the hyperspherical semantic space can cause noise in the similarity measures. We then propose to refine the initial query representation using a nearby example from the target domain, as illustrated in Figure~\ref{fig:refinement}.

We refine the query representation $p_0$ by performing a spherical linear interpolation with a relevant representation $p_1$. The refined representation $\hat{p}$ is:
\begin{equation}
\label{eq:refinement}
\hat{p}(p_{0},p_{1}|\lambda)={\frac {\sin {\big((1-\lambda)\Omega \big)}}{\sin \Omega }} p_{0} + {\frac {\sin\big(\lambda\Omega \big)}{\sin \Omega }}p_{1},
\end{equation}
where $\Omega = \arccos{(p_0 \cdot p_1)}$ and $\lambda \in [0,1]$ controls the amount of mixture in the refinement process. The higher the value of lambda is, the further away the refined representation is from the original representation $p_0$. Intuitively, the refinement performs a weighted signal averaging to reduce the noise present in the initial representation.
In retrieval, we set $p_1$ as the 1-nearest neighbour of $p_0$ in the target set. This mixture doesn't require any label and relies on the fact that the recall at one is usually very high.
In classification, $p_1$ is the word embedding of the category name.

\section{Open cross-domain visual search}\label{sec:open}
In the first set of experiments, we demonstrate the ability to perform open cross-domain visual search in three ways. We note that this is a new setting, making direct comparisons to existing works infeasible. First, we demonstrate how we can search from any source to any target domain without hassle. Second, we show the potential and positive effect of searching from multiple source domains for any target domain. Third, we exhibit the possibility of searching in multiple target domains simultaneously.

\paragraph{Setup}
We evaluate on the recently introduced \emph{DomainNet}~\cite{peng2018moment}, which contains 596,006 images from 345 classes. Images are gathered from six visual domains: \textit{clipart}, \textit{infograph}, \textit{painting}, \textit{pencil}, \textit{photo} and \textit{sketch}.
We consider retrieval in  \textit{zero}- and \textit{many}-shot evaluations:
(\textit{i}) in the \textit{zero}-shot evaluation, $\mathcal{Y}$ is split into $\mathcal{Y}_{train}$ and $\mathcal{Y}_{test}$, with $\mathcal{Y}_{train} \cap \mathcal{Y}_{test} = \emptyset$, \ie, categories to be searched during inference have not been seen during training;
(\textit{ii}) the \textit{many}-shot evaluation uses the same categories during both training and testing.
The zero-shot evaluation randomly splits samples into 300 training and 45 testing classes. Following the zero-shot learning good practices in Xian~\etal\cite{xian2018zero}, we have verified the presence of the 345 categories of \textit{DomainNet}~\cite{peng2018moment} in \textit{ImageNet}~\cite{ILSVRC15}, where we identify 188 separate categories. From this list of separate categories, we randomly sample 45 zero-shot categories with at least 40 samples per class in every domain.
The many-shot evaluation follows the original splits from Peng~\etal\cite{peng2018moment}.
We report the mean average precision (mAP@all).

\paragraph{Implementation details}
Throughout the paper and unless stated otherwise, we use SE-ResNet50~\cite{hu2018squeeze} pre-trained on ImageNet~\cite{ILSVRC15} as a backbone, and word2vec trained on a Google News corpus~\cite{mikolov2013efficient} as the common semantic space. We remove the final classifier layer of SE-ResNet50, and replace it with a fully-connected layer of size $D=300$ initialized with random weights. The new layer is followed by a linear activation and batch normalization~\cite{ioffe2015batch}. We optimize the loss in Equation~\ref{eq:loss} with Nesterov momentum~\cite{sutskever2013importance} by setting the coefficient to 0.9. We apply a learning rate of $1\mathrm{e}{-4}$ with cosine annealing without warm restarts~\cite{loshchilov2017sgdr} and a batch size of 128.
We use a scaling factor $s$ of 20, and decrease it to 10 for Sections~\ref{sec:sota2} and~\ref{sec:sota3}. We set $\lambda=0.7$ when evaluating on unseen classes (\ie zero-shot and few-shot evaluations) and to 0.4 when evaluating on seen classes (\ie many-shot evaluation).
The implementation rests on the Pytorch~\cite{paszke2019pytorch} framework and image similarities are computed with the Faiss~\cite{JDH17} library. Word embeddings of class names are extracted with the Gensim~\cite{rehurek_lrec} library.

\begin{table}[t]
\caption{\textbf{Visual search from sketches} as a source to any target domain comparison with SAKE~\cite{liu2019semantic} in mAP@all. Our formulation achieves competitive results in both \textit{zero}- and \textit{many}-shot evaluations.}
\centering
\resizebox{0.95\linewidth}{!}{
\begin{tabular}{lcccccc}
\toprule
target domain & \multicolumn{2}{c}{\textbf{\textit{zero}-shot}} & \multicolumn{2}{c}{\textbf{\textit{many}-shot}}\\
 & SAKE & \textit{This paper} & SAKE & \textit{This paper} \\
\cmidrule(lr){1-1}\cmidrule(lr){2-3}\cmidrule(lr){4-5}
clipart & 0.199 & \textbf{0.236} & 0.268 & \textbf{0.373} \\
infograph & 0.080 & \textbf{0.083} & 0.097 & \textbf{0.131} \\
painting & 0.118 & \textbf{0.142} & 0.203 & \textbf{0.317} \\
pencil & 0.181 & \textbf{0.214} & 0.230 & \textbf{0.328} \\
photo & 0.206 & \textbf{0.240} & 0.358 & \textbf{0.496} \\
\bottomrule
\end{tabular}
}
\label{tab:sake}
\vspace{-1em}
\end{table}

\subsection{From any source to any target domain}
First, we demonstrate how searching from any source to any target domain in an open setting is trivially enabled by our approach.
Figure~\ref{fig:any} shows the result of 72 cross-domain search evaluations; corresponding to all six cross-domain pairs for both zero- and many-shot evaluations. In our formulation, such an exhaustive evaluation is enabled by training only six models, one for every domain. For comparison, a domain adaptation approach --the standard in current cross-domain search methods-- requires a pair-wise training of all available domain combinations. Moreover, our formulation allows for an easy integration of new domains, as only the mapping from a new visual domain to the shared semantic space needs to be trained. While approaches based on pair-wise training scale with a quadratic complexity to the number of domains, we scale linearly.

In the zero-shot evaluation with an evaluation on the unseen classes (Figure~\ref{fig:anya}), the \textit{photograph} domain provides the most effective search whether used as source or target. One reason is the number of available images, which is up to four times larger than other domains. On the other hand, \textit{infographs} and \textit{sketches} are very diverse in terms of scale and visual representations, which induces a much more difficult search.

In the many-shot evaluation with an evaluation on all classes (Figure~\ref{fig:anyb}), the \textit{photograph} domain exhibits a similar behaviour. Though, in this case the search performance for \textit{sketches} is at the same level as other considered domains, such as \textit{clipart}, \textit{painting} or \textit{pencil}. Seeing all classes helps the prototype learner to better grasp the variability in \textit{sketches}. The \textit{infograph} domain remains the most challenging.
We conclude from the first demonstration that search from any source to any target domain is not only feasible with our approach, it can be done easily for both zero- and many-shot evaluations since we bypass the need to align different domains.

We quantitatively compare with the state-of-the-art SAKE~\cite{liu2019semantic} on zero-shot sketch-based image retrieval. We run SAKE from the original source code provided by the authors. Table~\ref{tab:sake} presents the results when considering sketches as the source domain and retrieving images in any of the other domains. SAKE has been proposed with a zero-shot evaluation design from the start, which makes it strong in this setting. Indeed, results are close, we only observe an improvement of 0.3\% (\textit{infograph}) up to 3.7\% (\textit{clipart}). When the evaluation focuses on a large number of categories, we notice higher gains from 3.4\% (\textit{infograph}) up to 13.8\% (\textit{photograph}) in the many-shot evaluation. Our embedding space is better partitioned for all categories thanks to the semantic prototypes. Overall, our formulation provides competitive performance in both zero- and many-shot evaluations with a simpler training.

\begin{figure}[t]
\vspace{-1em}
\centering
\subfloat[\textit{Zero}-shot evaluation]{
\includegraphics[width=0.48\linewidth]{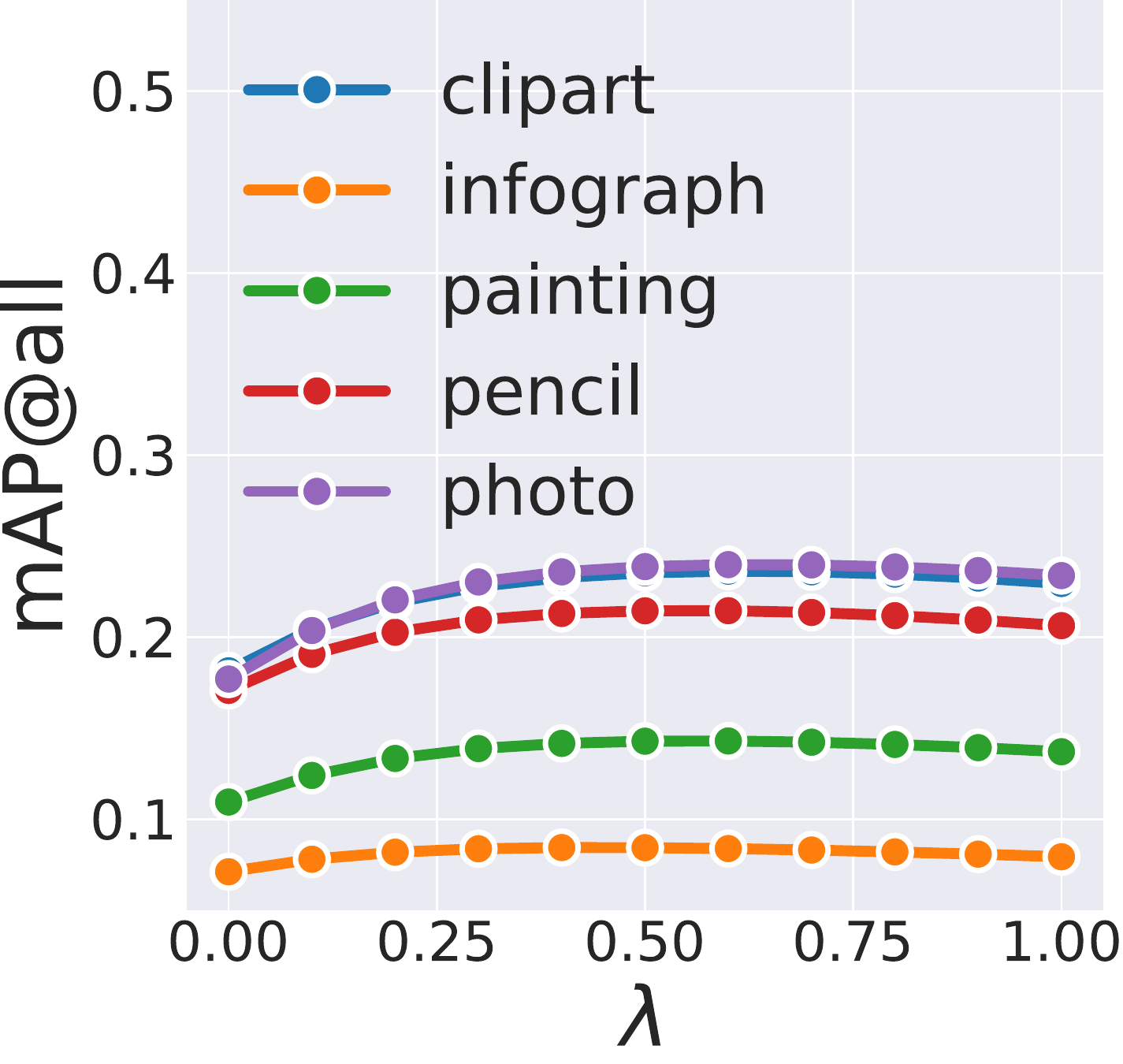}\label{fig:anyref:a}}\hfill
\subfloat[\textit{Many}-shot evaluation]{
\includegraphics[width=0.48\linewidth]{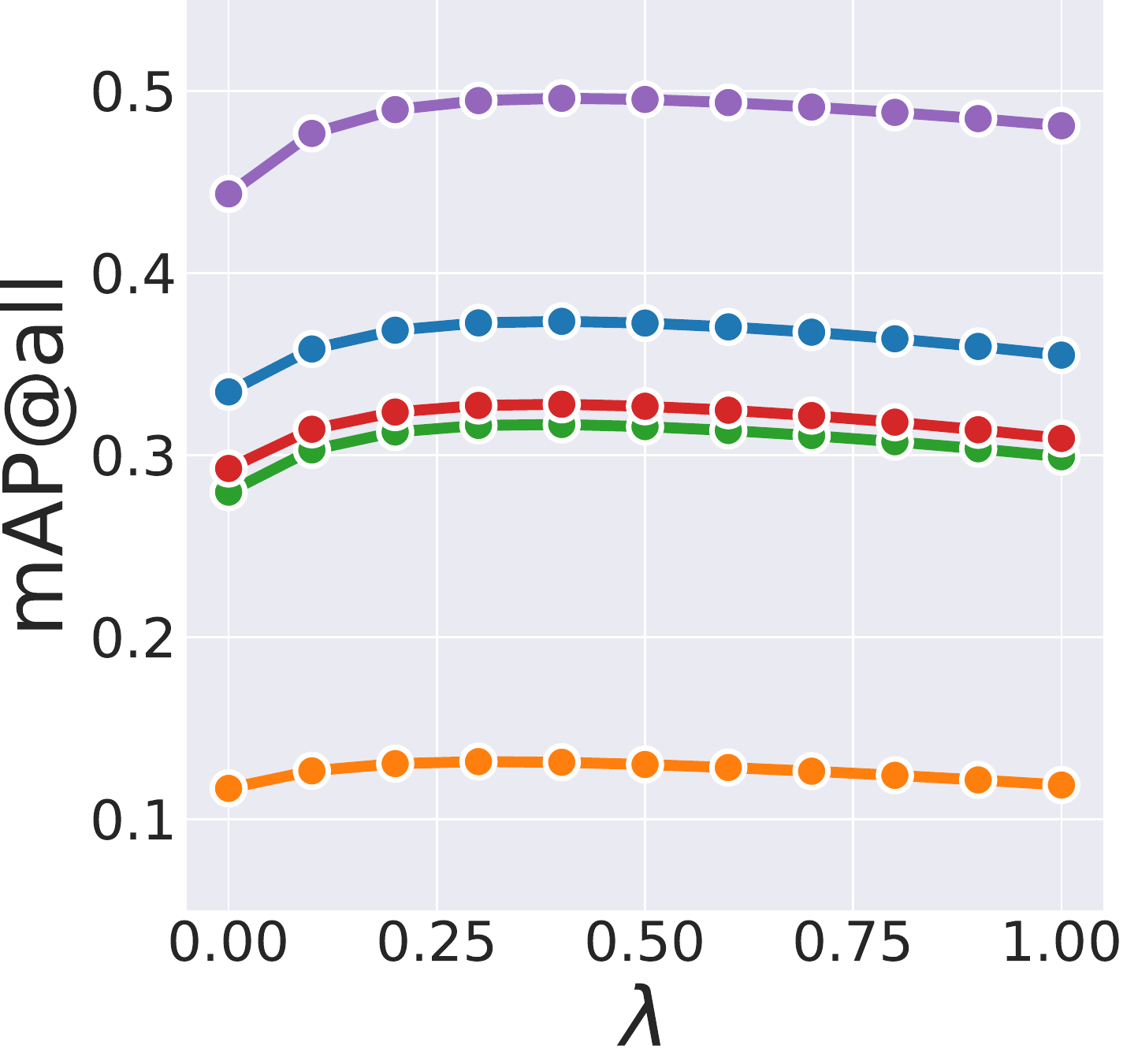}\label{fig:anyref:b}}
\caption{\textbf{Ablation on cross-domain query refinement} on DomainNet, with \textit{sketches} as a source. Refining the source representation always improves the retrieval performance.
\label{fig:anyref}}
\vspace{-1em}
\end{figure}

Finally, we also assess the importance of the proposed refinement module of Equation~\ref{eq:refinement}. Figure~\ref{fig:anyref} illustrates the effect of our cross-domain prototypical refinement when searching in any target domain from the \textit{sketch} domain. 
We create a mixture between the \textit{sketch} query ($\lambda=0$) and its nearest neighbour in the gallery ($\lambda=1$) for retrieval.
For both zero- and many-shot evaluations, refining the representations improves the performance. We observe a need for a lower mixture for the many-shot evaluation, as classes are all seen during training compared to the zero-shot evaluation.
Refining the representations helps to bridge the inherent cross-domain gap.

\begin{table}[t]
\caption{\textbf{Demonstration 2} for visual search from multiple sources to any target domain (absolute improvement in mAP@all). In our approach, searching from multiple sources is as easy as using a single source, as we only have to average their positions in the common semantic space. Searching (a) from multiple diverse domains is preferred when the source is less informative, while (b) more examples from the same domain are preferred when the source is more informative.}
\vspace{-0.5em}
\centering
\subfloat[{\small Improving the less informative sketch representations}\label{tab:two:a}]{
\resizebox{0.95\linewidth}{!}{
\begin{tabular}{lcccccc}
\toprule
target domain & \multicolumn{3}{c}{\textbf{\textit{zero}-shot}} & \multicolumn{3}{c}{\textbf{\textit{many}-shot}}\\
 & sk+sk & sk+in & sk+ph & sk+sk & sk+in & sk+ph \\
\cmidrule(lr){1-1}\cmidrule(lr){2-4}\cmidrule(lr){5-7}
clipart & +.057 & +.072 & +\textbf{.211} & +.097 & +.036 & +\textbf{.178} \\
infograph & +.018 & +.067 & +\textbf{.107} & +.031 & +.002 & +\textbf{.075} \\
painting & +.035 & +.080 & +\textbf{.186} & +.079 & +.029 & +\textbf{.154} \\
pencil & +.054 & +.060 & +\textbf{.154} & +.083 & +.043 & +\textbf{.156} \\
photo & +.064 & +.112 & +\textbf{.328} & +.127 & +.049 & +\textbf{.185} \\
\bottomrule
\end{tabular}
}}\\
\subfloat[{\small Improving the more informative photograph representations}\label{tab:two:b}]{
\resizebox{0.95\linewidth}{!}{
\begin{tabular}{lcccccc}
\toprule
target domain & \multicolumn{3}{c}{\textbf{\textit{zero}-shot}} & \multicolumn{3}{c}{\textbf{\textit{many}-shot}}\\
 & ph+ph & ph+in & ph+sk & ph+ph & ph+in & ph+sk \\
\cmidrule(lr){1-1}\cmidrule(lr){2-4}\cmidrule(lr){5-7}
clipart & \textbf{+.070} & +.012 & +.048 & \textbf{+.075} & +.002 & +.067 \\
infograph & \textbf{+.029} & -.035 & +.005 & \textbf{+.027} & -.062 & +.018 \\
painting & \textbf{+.052} & +.011 & +.008 & \textbf{+.061} & +.004 & +.049 \\
pencil & \textbf{+.054} & +.012 & +.037 & \textbf{+.066} & +.000 & +.057 \\
sketch & +.041 & +.001 & \textbf{+.202} & +\textbf{.075} & -.013 & -.030 \\
\bottomrule
\end{tabular}
}}
\label{tab:demo2}
\vspace{-1em}
\end{table}

\subsection{From multiple sources to any target domain}
Second, we demonstrate the potential to search from multiple source domains. Due to the generic nature of our approach, we are not restricted to search from a single source. We show that a multi-source search benefits the search in any target domain. 

For this experiment, we start from the \textit{sketch} domain as a source and investigate the effect of including queries from the most effective source (\textit{photographs}) and the least effective source (\textit{infographs}). Table~\ref{tab:two:a} highlights the positive effect of searching with an additional domain, rather than a single source domain. When using multiple sources, we simply average the positions in the common semantic space. For fairness, we also evaluate search using two \textit{sketches}. Across all settings, we find that searching from multiple queries improves relative to using one single \textit{sketch} query. In the zero-shot evaluation, including \textit{infographs} and \textit{photographs} improves upon sketch-based search only. In the many-shot evaluation, including \textit{infographs} improves upon search by one \textit{sketch}, but not by two \textit{sketches}, which is not surprising given the low scores for infographs individually. \textit{Photographs} with \textit{sketches} obtain the highest scores, regardless of the target domain or the evaluation setting.

We also consider a more challenging multi-source search scenario where we search from the most informative source (\textit{photograph}) and one of the least informative sources (\textit{infograph} or \textit{sketch}). Table~\ref{tab:two:b} confirms the positive effect of searching with an additional domain. Adding \textit{infographs} only improves the results marginally. Performance can even decrease when searching within one of the least informative domains, because the combination creates a destructive noise that moves the initial representation to a wrong direction. Adding \textit{sketches} can benefit searching within \textit{sketches} when the uncertainty is high, as in a zero-shot evaluation, but slightly decreases the score when the uncertainty is low, as in a many-shot evaluation. In the other target domains, \textit{sketches} are much more effective than \textit{infographs} when added to \textit{photographs}. Though, the improvement is lower than searching from two \textit{photographs}. When searching from an informative source domain, combining it with itself improves more than a combination with a less informative domain for both zero- and many-shot evaluations.

This demonstration shows the potential of searching from multiple sources. It is better to diversify the search by using multiple diverse domains when the source is less informative while more queries from the same domain are preferred when the source is more informative. Similar to the first demonstration, this evaluation is a trivial extension to our approach, as we only have to average positions in the shared semantic space, regardless of the domain the examples come from.

\begin{figure*}[t]
\vspace{-1em}
\centering
\hfill
\begin{overpic}[width=0.45\linewidth]{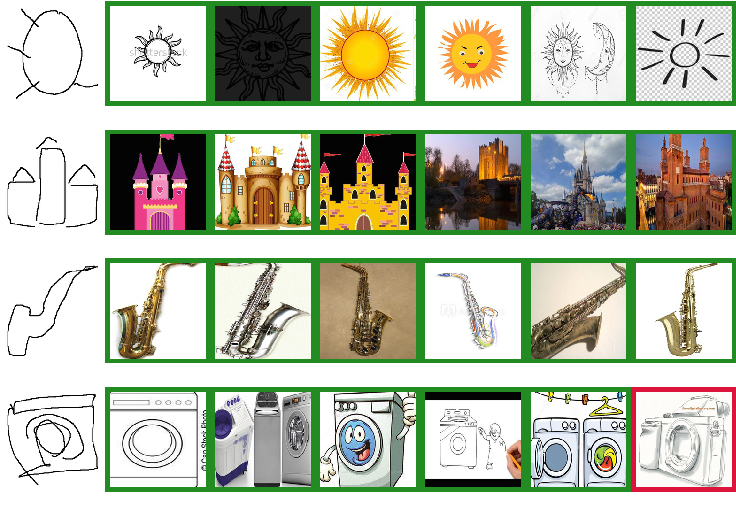}
\put(-2,60){\scriptsize \rotatebox{90}{sun}}
\put(-2,43){\scriptsize \rotatebox{90}{castle}}
\put(-2,21){\scriptsize \rotatebox{90}{saxophone}}
\put(-2,5){\scriptsize \rotatebox{90}{machine}}
\put(-5,5){\scriptsize \rotatebox{90}{washing}}

\put(16,53){\scriptsize \textit{pencil}}
\put(16,36){\scriptsize \textit{clipart}}
\put(16,18){\scriptsize \textit{photo}}
\put(16,1){\scriptsize \textit{clipart}}

\put(30,53){\scriptsize \textit{clipart}}
\put(30,36){\scriptsize \textit{clipart}}
\put(30,18){\scriptsize \textit{photo}}
\put(30,1){\scriptsize \textit{photo}}

\put(44,53){\scriptsize \textit{clipart}}
\put(44,36){\scriptsize \textit{clipart}}
\put(44,18){\scriptsize \textit{photo}}
\put(44,1){\scriptsize \textit{clipart}}

\put(58,53){\scriptsize \textit{clipart}}
\put(58,36){\scriptsize \textit{photo}}
\put(58,18){\scriptsize \textit{pencil}}
\put(58,1){\scriptsize \textit{pencil}}

\put(72,53){\scriptsize \textit{pencil}}
\put(72,36){\scriptsize \textit{painting}}
\put(72,18){\scriptsize \textit{photo}}
\put(72,1){\scriptsize \textit{clipart}}

\put(86,53){\scriptsize \textit{pencil}}
\put(86,36){\scriptsize \textit{photo}}
\put(86,18){\scriptsize \textit{photo}}
\put(86,1){\scriptsize \textit{pencil}}
\end{overpic}\hfill
\begin{overpic}[width=0.45\linewidth]{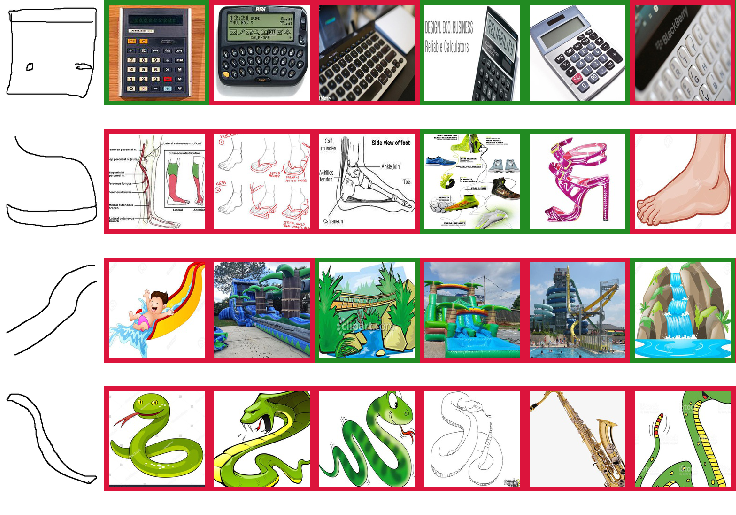}
\put(-2,57){\scriptsize \rotatebox{90}{calculator}}
\put(-2,43){\scriptsize \rotatebox{90}{shoe}}
\put(-2,25){\scriptsize \rotatebox{90}{river}}
\put(-2,5){\scriptsize \rotatebox{90}{waterslide}}

\put(16,53){\scriptsize \textit{photo}}
\put(16,36){\scriptsize \textit{photo}}
\put(16,18){\scriptsize \textit{clipart}}
\put(16,1){\scriptsize \textit{clipart}}

\put(30,53){\scriptsize \textit{photo}}
\put(30,36){\scriptsize \textit{pencil}}
\put(30,18){\scriptsize \textit{clipart}}
\put(30,1){\scriptsize \textit{clipart}}

\put(44,53){\scriptsize \textit{photo}}
\put(44,36){\scriptsize \textit{photo}}
\put(44,18){\scriptsize \textit{clipart}}
\put(44,1){\scriptsize \textit{clipart}}

\put(58,53){\scriptsize \textit{photo}}
\put(58,36){\scriptsize \textit{infograph}}
\put(58,18){\scriptsize \textit{pencil}}
\put(58,1){\scriptsize \textit{pencil}}

\put(72,53){\scriptsize \textit{photo}}
\put(72,36){\scriptsize \textit{pencil}}
\put(72,18){\scriptsize \textit{photo}}
\put(72,1){\scriptsize \textit{photo}}

\put(86,53){\scriptsize \textit{photo}}
\put(86,36){\scriptsize \textit{photo}}
\put(86,18){\scriptsize \textit{clipart}}
\put(86,1){\scriptsize \textit{clipart}}
\end{overpic}\hfill\null
\caption{\textbf{Demonstration 3} for visual search from any source to multiple target domains. Correct results are in \textit{{\color{ForestGreen}green}}, incorrect in \textit{{\color{BrickRed}red}}.
For abstract categories such as ``sun'', abstract domains such as \textit{clipart} or \textit{pencil} drawings tend to be retrieved first. When \textit{sketches} are more ambiguous such as ``shoe'', some retrieved results are incorrect but resemble the shape.
}
\label{fig:demo3}
\vspace{-1em}
\end{figure*}

\begin{figure}[t]
\centering
\includegraphics[width=0.7\linewidth]{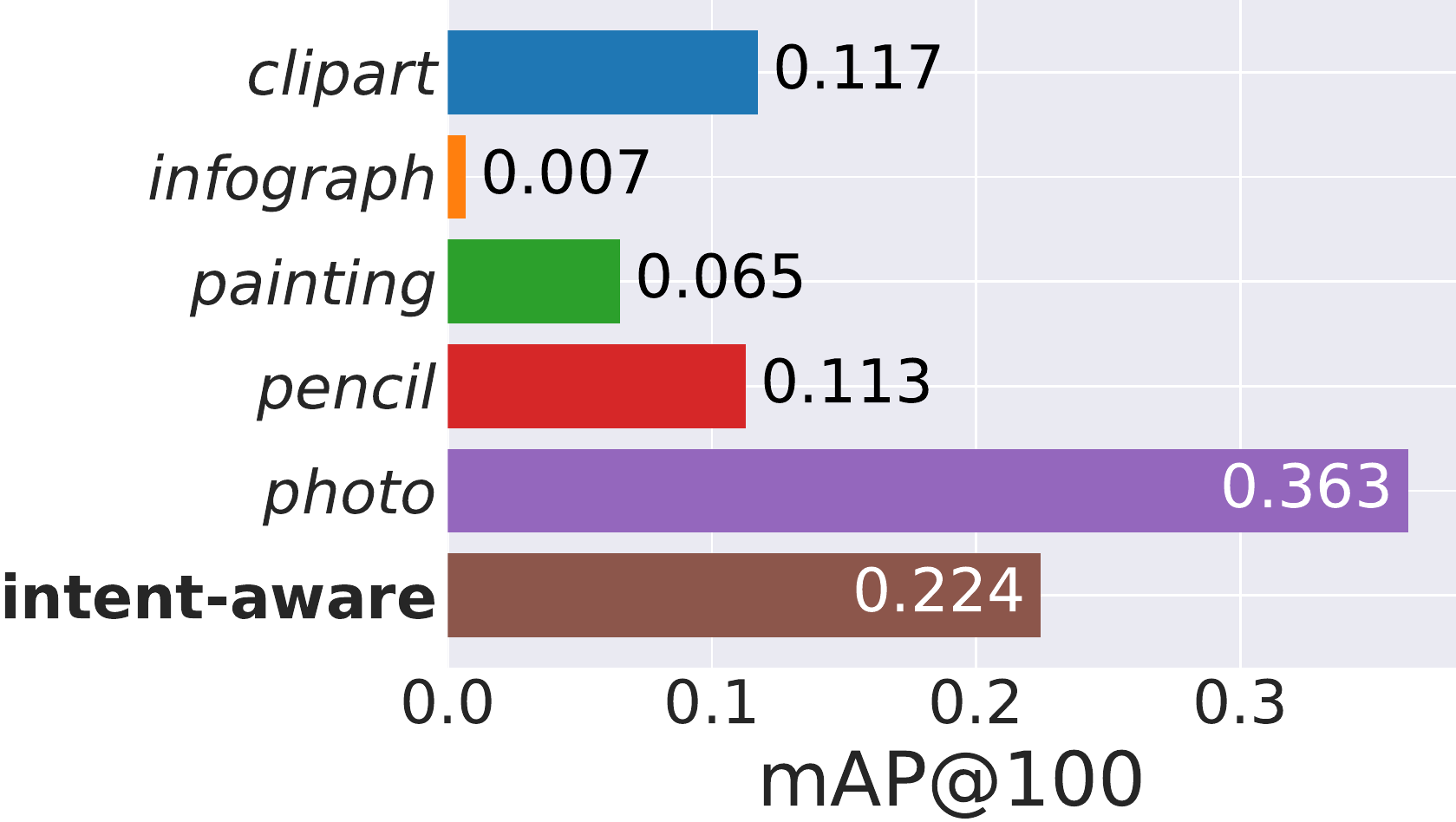}
\caption{\textbf{Intent-aware evaluation} for visual search from sketches to the other five target domains. Correct retrieved images in the top-ranked results more likely come from the \textit{photograph} than the \textit{infograph} domain.}
\label{fig:intent}
\vspace{-1em}
\end{figure}

\subsection{From any source to multiple target domains}
Third, we demonstrate our ability to search in multiple domains simultaneously.
This setting has potential applications for example in untargeted portfolio browsing, where a user may want to explore all possible visual expressions of a category.
Exploring in multiple domains also highlights whether certain categories have a preference towards specific domains, which offers an insight on how to best depict those categories.
Note that this setting can also be easily extended to include also multiple domains as a source.
For the sake of clarity, we use \textit{sketch} as the source domain and search in the other five domains in a many-shot evaluation.

Figure~\ref{fig:demo3} provides qualitative results for eight \textit{sketches} from different categories. We first observe that the results come from multiple target domains, without being explicitly told to do so. We do not need to align results from different target domains, since we measure distance in the common semantic space. For categories such as \say{sun}, we have a bias towards retrieving abstract depictions, such as \textit{pencil} drawings and \textit{cliparts}, as the \say{sun} is a category with a clear abstract representation. \say{Castle} on the other hand has a bias towards both distinct \textit{cliparts}, as well as \textit{photographs} and \textit{paintings}. In both cases, all top results are relevant. For categories with more ambiguous \textit{sketches}, such as \say{river} or \say{calculator}, retrieved examples resemble the shape of the provided \textit{sketch}, but do not match the category. Overall, we conclude that searching in multiple domains is not only trivial in our approach, but is also an indicator of the presence of preferential domains for depicting categories.

We also quantitatively measure the retrieval performance when searching from sketches to the other five target domains simultaneously. When computing the mAP@100, we obtain a score of 0.565. Though, this measure does not take into account the differences and diversity among domains, as it considers all of them as similar. As such, we report the intent-aware mAP~\cite{agrawal2009diversifying}. Extending the mAP to an intent-aware formulation provides an estimate of the result diversity by: (i) computing the mAP per domain, and (ii) summing them with a weighting that corresponds to the occurrences of every category within each domain. Figure~\ref{fig:intent} shows the per domain and intent-aware mAP@100. The \textit{photograph}-mAP@100 is the highest score, which indicates correct \textit{photographs} are in the top-ranked results compared with other target domains. The \textit{infograph}-mAP@100 obtains the lowest score, which means that there are very few correct \textit{infographs} in the top-ranked results. When the differences among domains are taken into consideration, the intent-aware mAP@100 results in 0.224. In a search within multiple domains, the informativeness of each domain influences the top-ranked results.

\section{Closed cross-domain visual search}\label{sec:close}
Our approach is geared towards open cross-domain visual search, as demonstrated in the previous section. To get insight in the effectiveness of our approach for cross-domain visual search in general, we also perform an extensive comparative evaluation on standard cross-domain settings, which search between two domains. In total, we compare on three of the most popular cross-domain search tasks, namely zero-shot sketch-based image retrieval~\cite{sketchy2016,shen2018zero}, few-shot sketch-based image classification~\cite{hu2018sketch}, and many-shot sketch-based 3D shape retrieval~\cite{li2013shrec,li2014shrec}.
For our approach, we simply train one mapping function for the source domain, and one for the target domain using the examples provided during training.
Below, we present each comparison separately.

\subsection{Zero-shot sketch-based image retrieval}\label{sec:sota1}
\paragraph{Setup} Zero-shot sketch-based image retrieval focuses on retrieving natural images (target domain) from a sketch query (source domain). We evaluate on two datasets.
\emph{TU-Berlin Extended}~\cite{eitz2012hdhso,zhang2016sketchnet} contains 20,000 sketches and 204,070 images from 250 classes. Following Shen~\etal~\cite{shen2018zero}, we select 220 classes for training and 30 classes for testing.
\emph{Sketchy Extended}~\cite{sketchy2016,liu2017deep} contains 75,481 sketches and 73,002 images from 125 classes. Similarly, following Shen~\etal~\cite{shen2018zero}, we select 100 classes for training and 25 classes for testing.
For fair comparison with Liu~\etal\cite{liu2019semantic}, we select the same unseen classes for both datasets. Following recent works~\cite{shen2018zero,dutta2019semantically,liu2019semantic}, we report the mAP@all and the precision at 100 (prec@100) scores.

\begin{table}[t]
\caption{\textbf{Comparison 1} to zero-shot sketch-based image retrieval on TU-Berlin Extended and Sketchy Extended.
Aligning solely the semantics improves cross-domain image retrieval.
}
\centering
\vspace{-0.5em}
\subfloat[{\small Real-valued representations}\label{tab:zero:real}]{
\resizebox{0.95\linewidth}{!}{
\begin{tabular}{lcccc}
\toprule
 & \multicolumn{2}{c}{\textbf{TU-Berlin Extended}} & \multicolumn{2}{c}{\textbf{Sketchy Extended}}\\
 & mAP@all & prec@100 & mAP@all & prec@100\\
 \cmidrule(lr){1-1}\cmidrule(lr){2-3}\cmidrule(lr){4-5}
 EMS~\cite{lu2018learning}   & 0.259 & 0.369 & \deemph{n/a} & \deemph{n/a}\\
 CAAE~\cite{kiran2018zero}  & \deemph{n/a} & \deemph{n/a} & 0.196 & 0.284\\
 ADS~\cite{dey2019doodle}  & 0.110 & \deemph{n/a} & 0.369 & \deemph{n/a}\\
 SEM-PCYC~\cite{dutta2019semantically}   & 0.297 & 0.426 & 0.349 & 0.463\\
 SG~\cite{dutta2019style}   & 0.254 & 0.355 & 0.376 & 0.484\\
 SAKE~\cite{liu2019semantic}   & 0.475 & \textbf{0.599} & 0.547 & 0.692\\
\rowcolor{Gray}
\emph{This paper}   & \textbf{0.517} & 0.557 & \textbf{0.649} & \textbf{0.708}\\
\bottomrule
\end{tabular}
}}\\
\subfloat[{\small Binary representations}\label{tab:zero:bin}]{
\resizebox{0.95\linewidth}{!}{
\begin{tabular}{lcccc}
\toprule
  & \multicolumn{2}{c}{\textbf{TU-Berlin Extended}} & \multicolumn{2}{c}{\textbf{Sketchy Extended}}\\
 &  mAP@all & prec@100 & mAP@all & prec@100\\
\cmidrule(lr){1-1}\cmidrule(lr){2-3}\cmidrule(lr){4-5}
 EMS~\cite{lu2018learning}  & 0.165 & 0.252 & \deemph{n/a} & \deemph{n/a}\\
 ZSIH~\cite{shen2018zero}   & 0.220 & 0.291 & 0.254 & 0.340 \\
 SEM-PCYC~\cite{dutta2019semantically}  & 0.293 & 0.392 & 0.344 & 0.399 \\
 SAKE~\cite{liu2019semantic}  & 0.359 & 0.481 & 0.364 & 0.487 \\
\rowcolor{Gray}
 \emph{This paper}  & \textbf{0.404} & \textbf{0.517} & \textbf{0.466} & \textbf{0.618} \\
 \bottomrule
\end{tabular}
}}\\
\subfloat[{\small Generalized setting}\label{tab:zero:gzs}]{
\resizebox{0.95\linewidth}{!}{
\begin{tabular}{lcccc}
\toprule
  & \multicolumn{2}{c}{\textbf{TU-Berlin Extended}} & \multicolumn{2}{c}{\textbf{Sketchy Extended}}\\
 & mAP@all & prec@100 & mAP@all & prec@100\\
 \cmidrule(lr){1-1}\cmidrule(lr){2-3}\cmidrule(lr){4-5}
 ZSIH~\cite{shen2018zero}  & 0.142 & 0.218 & 0.219 & 0.296\\
 SEM-PCYC~\cite{dutta2019semantically}  & 0.192 & \textbf{0.298} & 0.307 & 0.364\\
 SG~\cite{dutta2019style} & 0.149 & 0.226 & 0.331 & 0.381 \\
\rowcolor{Gray}
\emph{This paper}   & \textbf{0.211} & 0.224 & \textbf{0.397} & \textbf{0.421} \\
\bottomrule
\end{tabular}
}}
\vspace{-1em}
\label{tab:sota1}
\end{table}

\begin{figure*}[t]
\vspace{-1em}
\centering
\hfill
\begin{overpic}[width=0.45\linewidth]{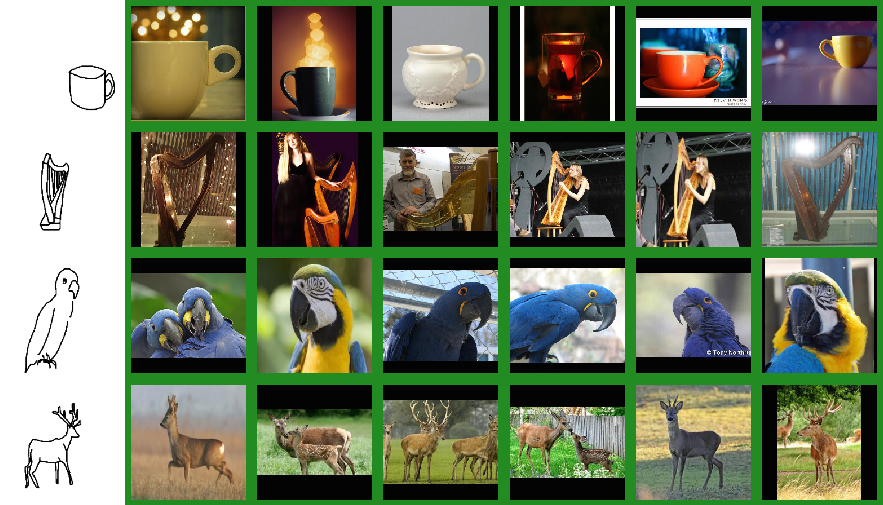}
\put(-2,48){\scriptsize \rotatebox{90}{\textit{cup}}}
\put(-2,33){\scriptsize \rotatebox{90}{\textit{harp}}}
\put(-2,18){\scriptsize \rotatebox{90}{\textit{parrot}}}
\put(-2,5){\scriptsize \rotatebox{90}{\textit{deer}}}
\end{overpic}\hfill
\begin{overpic}[width=0.45\linewidth]{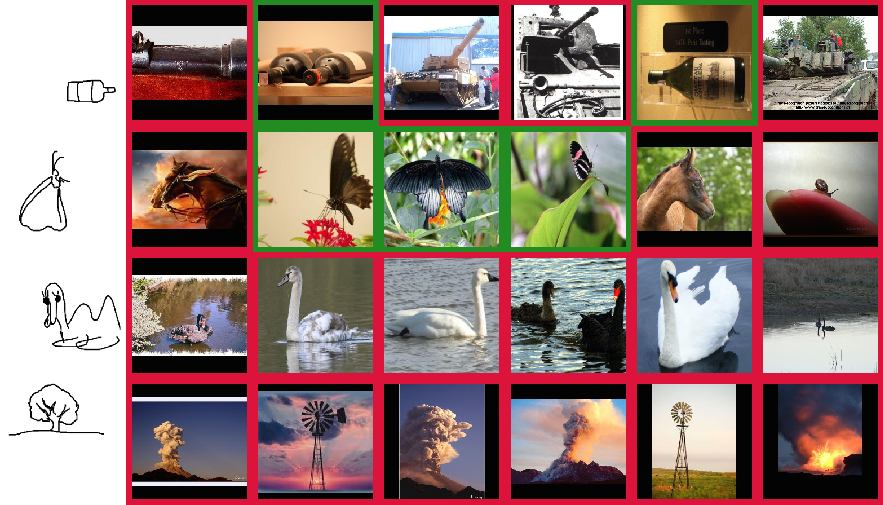}
\put(-2,48){\scriptsize \rotatebox{90}{\textit{bottle}}}
\put(-2,31){\scriptsize \rotatebox{90}{\textit{butterfly}}}
\put(-2,18){\scriptsize \rotatebox{90}{\textit{swan}}}
\put(-2,5){\scriptsize \rotatebox{90}{\textit{tree}}}
\end{overpic}\hfill\null
\caption{\textbf{Qualitative analysis} of zero-shot sketch-based image retrieval. We show eight sketches of Sketchy Extended, with correct retrievals in \textit{{\color{ForestGreen}green}}, incorrect in \textit{{\color{BrickRed}red}}. For typical sketches (\eg, ``cup''), the closest images are from the same category. For ambiguous sketches (\eg, ``tree'') or non-canonical views (\eg, ``butterfly''), our approach struggles.}
\label{fig:sota1-q}
\end{figure*}
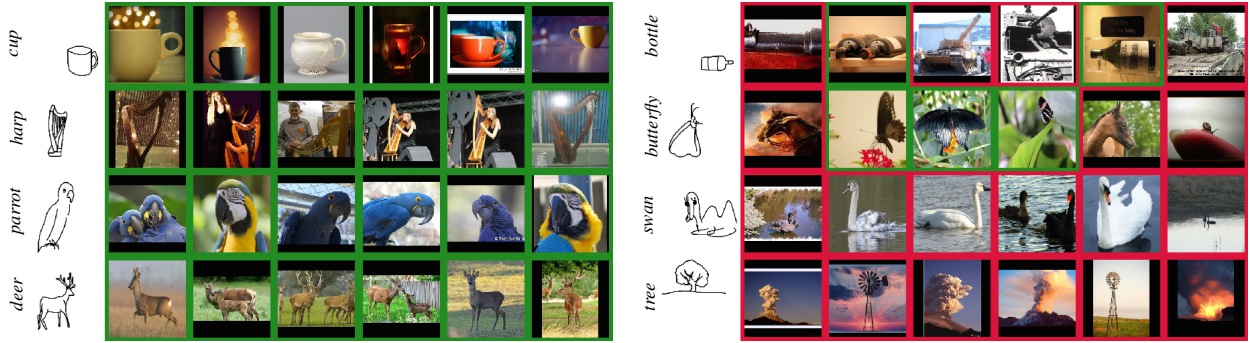

\paragraph{Results} Table~\ref{tab:zero:real} compares to six state-of-the-art baselines on both datasets.
Baselines mostly focus on bridging the domain gap between sketches and natural images with domain adaptation losses~\cite{ganin2016domain,gonzalez2018image}.
On Sketchy Extended, our approach outperforms other baselines. On TU-Berlin Extended, we obtain the highest mAP@all score, while the recently introduced SAKE by Liu~\etal\cite{liu2019semantic} obtains a higher prec@100 score. SAKE is better at grouping images from the same category together thanks to the preservation module that produces tightly distributed representations. Our method is better at retrieving relevant images in the first ranks as the refinement module reduces the noise in the query representations.

Following previous work in zero-shot sketch-based image retrieval~\cite{lu2018learning,shen2018zero,dutta2019semantically,liu2019semantic}, we also report the retrieval performance on binary representations.
As previously proposed in~\cite{dutta2019semantically,liu2019semantic}, real-valued representations are projected to a low-dimensional space and quantized with iterative quantization~\cite{gong2012iterative}. We compute the transformation on the training set and apply it on both sketch and image testing sets. Note that we first refine the representations, then apply iterative quantization.
Table~\ref{tab:zero:bin} compares the proposed formulation with binary representations of 64 dimensions. Compared to real-valued representations in Table~\ref{tab:zero:real}, we notice a higher drop in the mAP@all score compared to prec@100 score. Compared to other baselines, our semantic space based on word embeddings better preserves the information when compressed to a low-dimensional space.

As recently introduced by Dutta and Akata~\cite{dutta2019semantically}, we also evaluate on a generalized setting in Table~\ref{tab:zero:gzs}, where the gallery set also includes images from seen classes. Following their protocol, we reserve 20\% of the samples from the seen classes for evaluation and use VGG16~\cite{simonyan2014very} in this experiment for fair comparison.
On Sketchy Extended, our approach also outperforms other baselines. On TU-Berlin Extended, we obtain the highest mAP@all score, while SEM-PCYC by Dutta and Akata~\cite{dutta2019semantically} obtains a higher prec@100 score. Similar to the zero-shot evaluation, our method is better at ranking images than grouping them together.
Overall, focusing solely on semantic alignment outperforms alternatives on domain adaption or knowledge preservation across three different settings derived from two datasets.

\begin{figure}[t]
\centering
\hfill
\subfloat[Sketchy Extended]{
\includegraphics[width=0.42\linewidth]{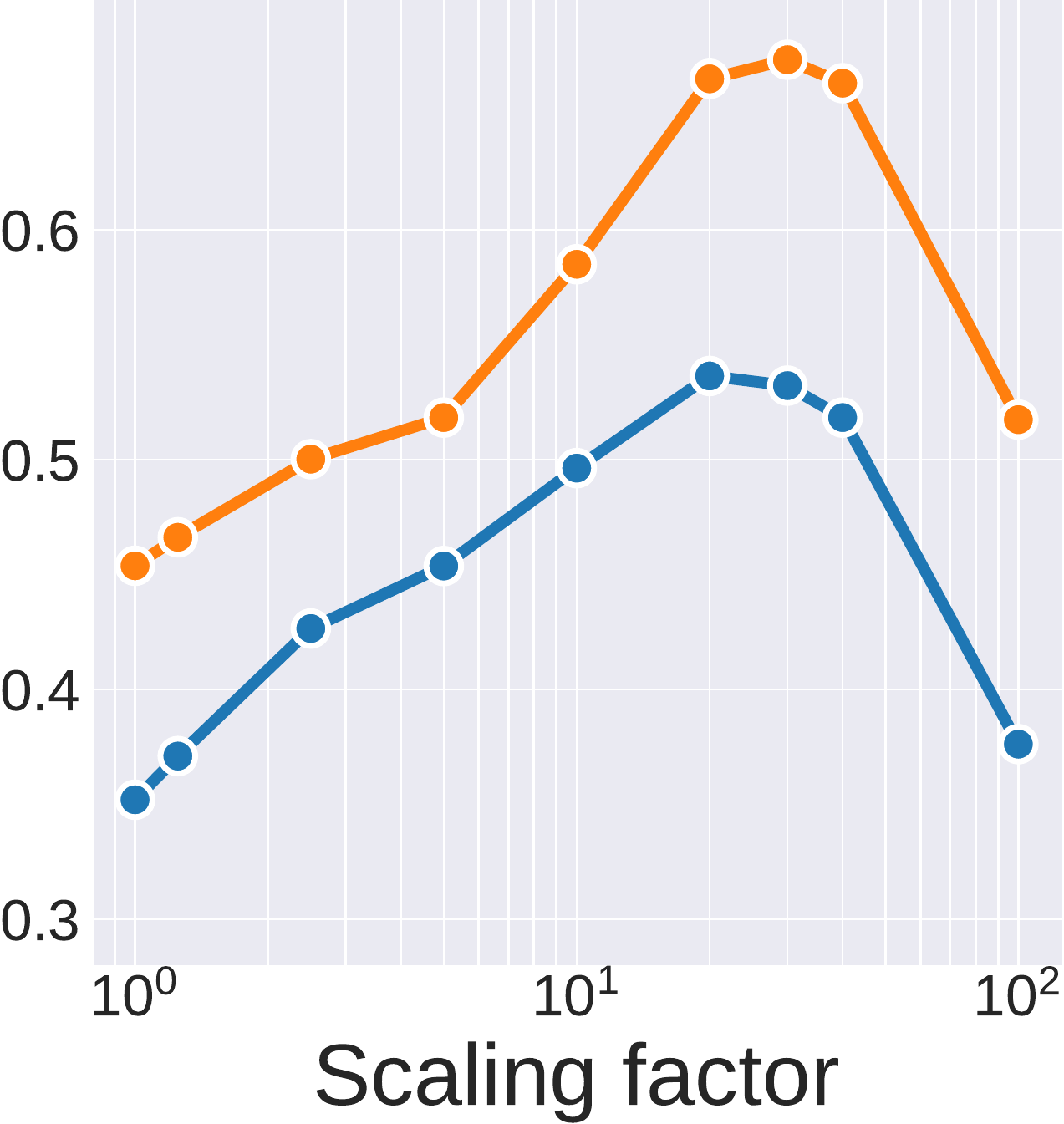}\label{fig:scaling:a}}\hfill
\subfloat[TU-Berlin Extended]{
\includegraphics[width=0.42\linewidth]{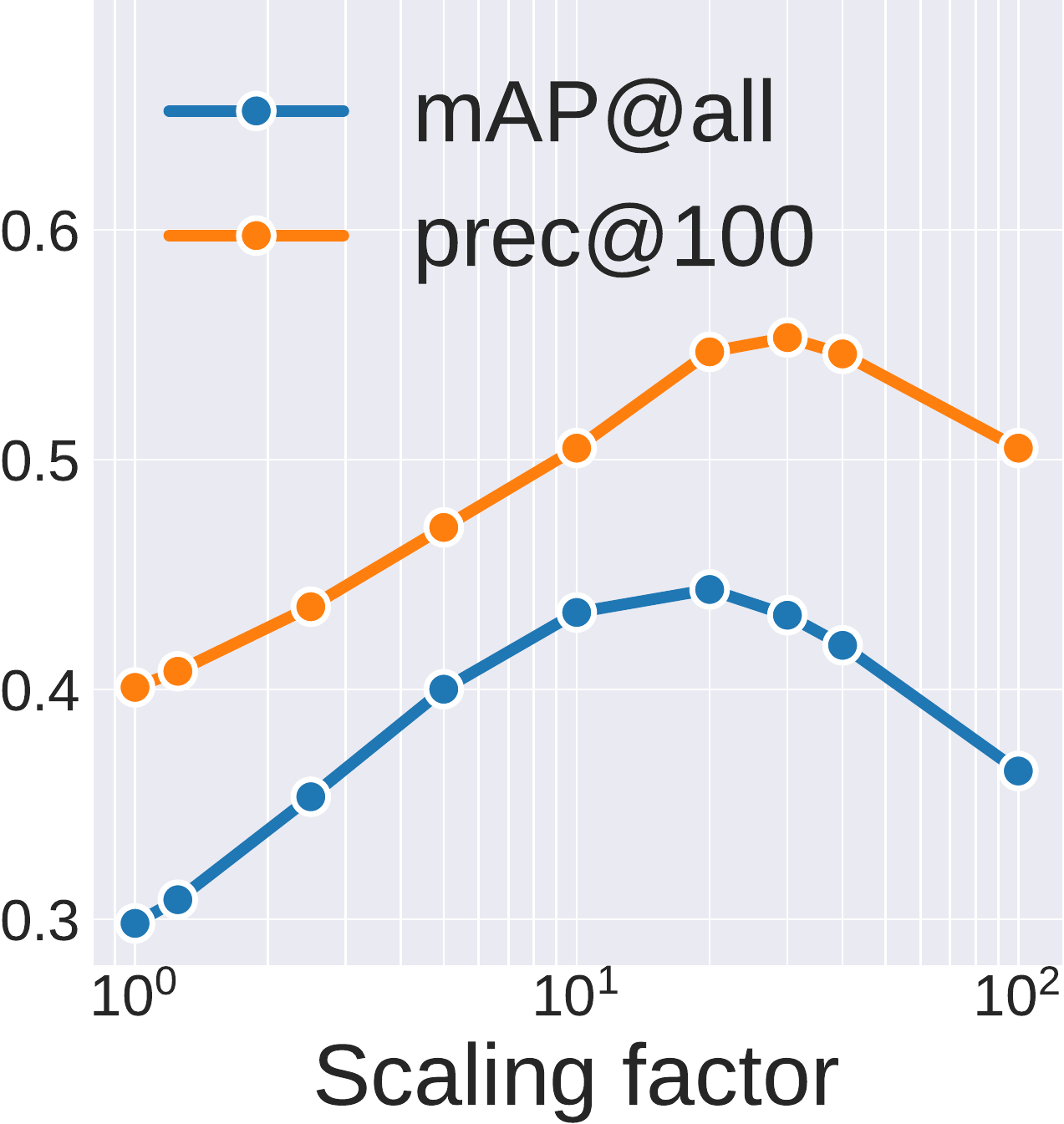}\label{fig:scaling:b}}\hfill\null
\caption{\textbf{Scaling hyper-parameter} ablation. We evaluate the scaling of the softmax function. $s=20$ yields the best results for both datasets, especially for the mAP@all score.}
\label{fig:scaling}
\vspace{-1em}
\end{figure}

To understand the effect of the distance scaling hyper-parameter defined in Equation~\ref{eq:py}, we vary its value on both datasets in Figure~\ref{fig:scaling}.
We observe the same behaviour on both datasets. When $s=1$ as in a common softmax function, it yields the lowest results. A higher scaling helps to narrow the probability distribution, resulting in a better retrieval performance. There is a tipping point around $s=20$, after which performance decreases.
Calibrating the softmax with a high distance scaling factor improves the retrieval performance.

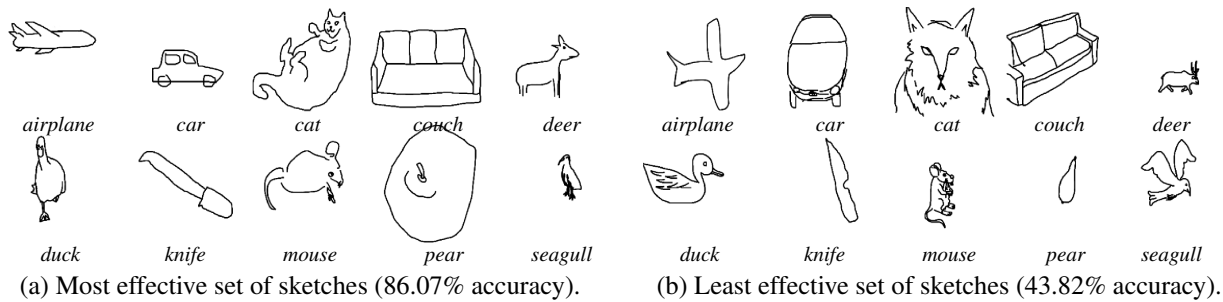
\begin{figure*}[t]
\vspace{-1em}
\centering
\hfill
\subfloat[][\centering Most effective set of sketches (86.07\% accuracy).]{
\begin{overpic}[width=0.45\linewidth, trim={30px 5px 15px 10px}, clip]{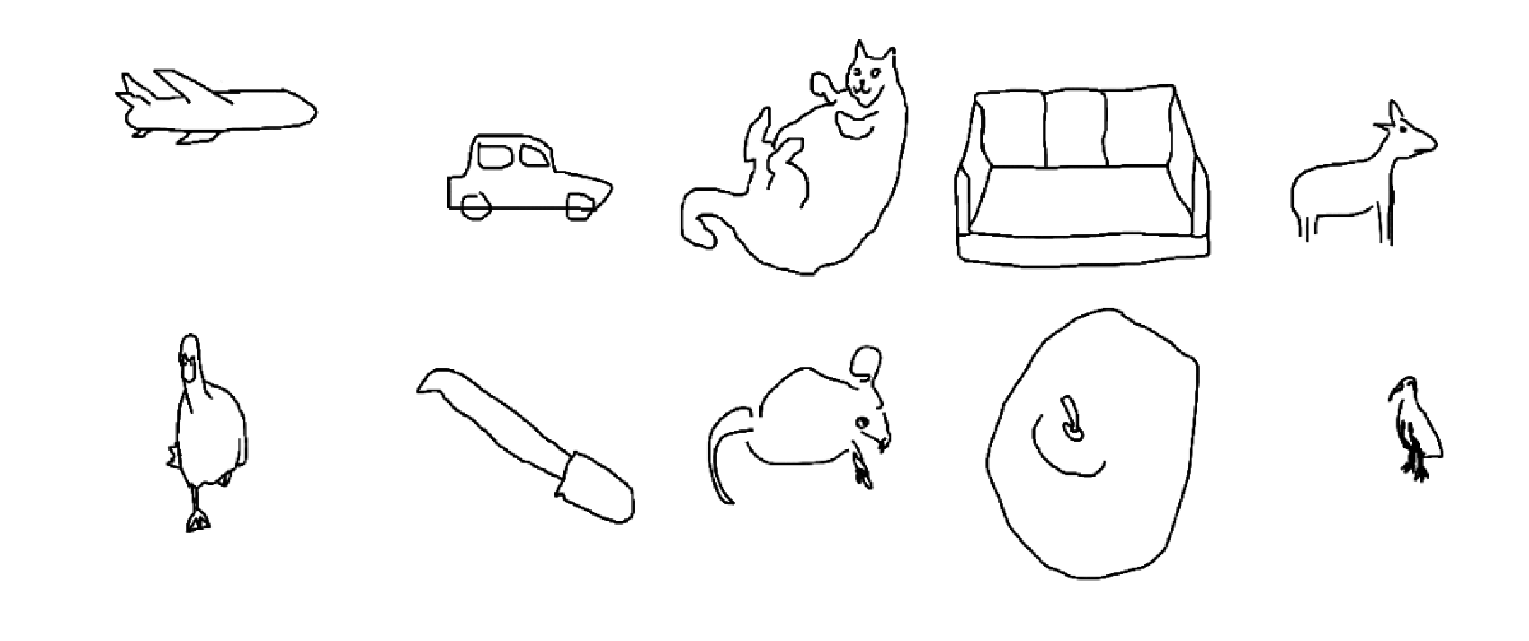}\label{fig:few:a}
\put(2,22){\footnotesize \textit{airplane}}
\put(28,22){\footnotesize \textit{car}}
\put(48,22){\footnotesize \textit{cat}}
\put(68,22){\footnotesize \textit{couch}}
\put(90,22){\footnotesize \textit{deer}}

\put(5,0){\footnotesize \textit{duck}}
\put(26,0){\footnotesize \textit{knife}}
\put(46,0){\footnotesize \textit{mouse}}
\put(70,0){\footnotesize \textit{pear}}
\put(88,0){\footnotesize \textit{seagull}}
\end{overpic}
}
\hfill
\subfloat[][\centering Least effective set of sketches (43.82\% accuracy).]{
\begin{overpic}[width=0.45\linewidth, trim={30px 5px 15px 10px}, clip]{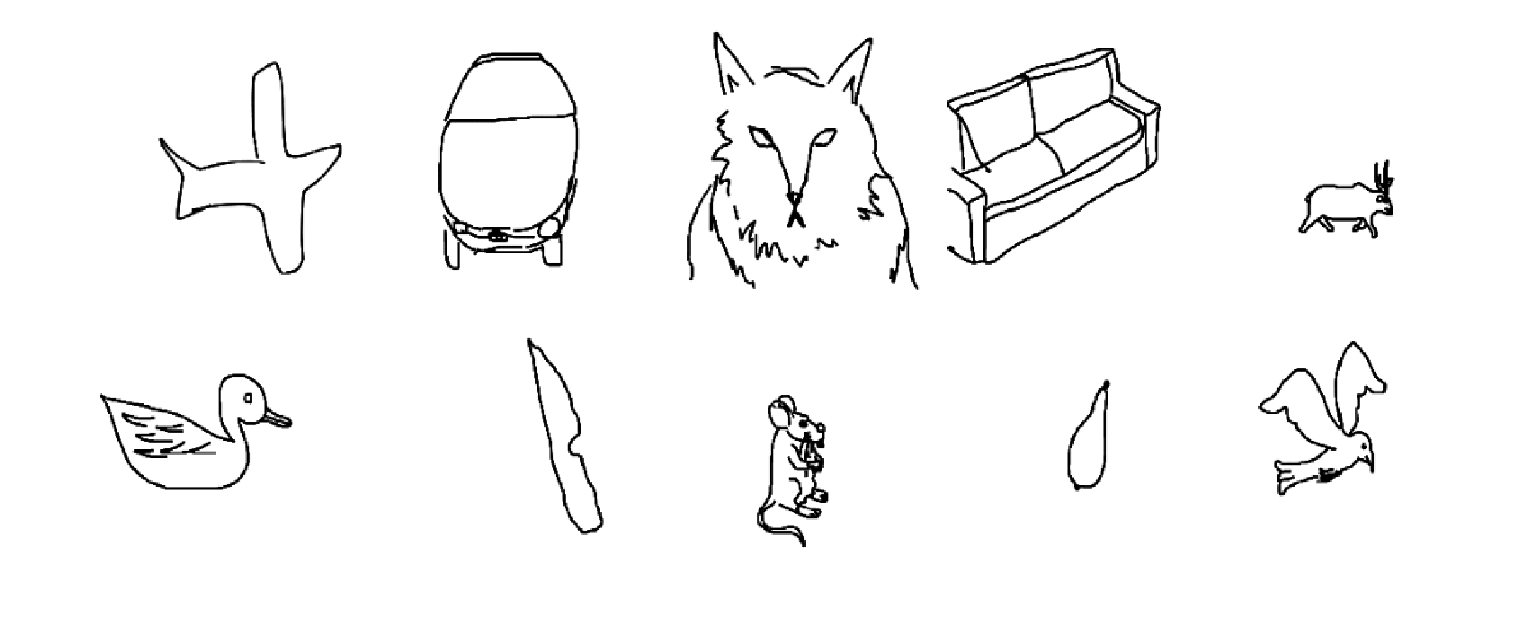}\label{fig:few:b}
\put(2,22){\footnotesize \textit{airplane}}
\put(28,22){\footnotesize \textit{car}}
\put(48,22){\footnotesize \textit{cat}}
\put(65,22){\footnotesize \textit{couch}}
\put(85,22){\footnotesize \textit{deer}}

\put(5,0){\footnotesize \textit{duck}}
\put(26,0){\footnotesize \textit{knife}}
\put(46,0){\footnotesize \textit{mouse}}
\put(67,0){\footnotesize \textit{pear}}
\put(83,0){\footnotesize \textit{seagull}}
\end{overpic}
}
\hfill\null
\caption{ \textbf{Qualitative analysis} of few-shot sketch-based image classification on a subsampled Sketchy Extended. (a) Since our approach condenses examples of category to a single prototype in the shared space, we obtain high scores when source sketches are detailed and in canonical views (\eg, ``deer'' or ``couch''). (b) The accuracy decreases when sketches are drawn badly (\eg, ``airplane''), or in non-canonical views (\eg, ``car'' or ``cat'').}
\label{fig:sota2-q}
\vspace{-1em}
\end{figure*}

\paragraph{Qualitative analysis} To understand which sketches trigger the performance of natural image retrieval, we provide several qualitative sketch queries with their top retrieved images in Figure~\ref{fig:sota1-q}.
Our approach works well for typical sketches of categories. For example, the \say{cup} or \say{parrot} sketches exhibit a typical definition of their respective categories. In return, the  search is very effective despite the variation in image appearance and viewpoints.
Results degrade when sketches are ambiguous or in non-canonical views. For example, the \say{tree} sketch can easily be confused with the smoke ring of a \say{volcano} or the shape of a \say{windmill}. Typical shape drawings of sketches matter for zero-shot image retrieval.

\begin{table}[t]
\caption{\textbf{Comparison 2} to few-shot sketch-based image classification on a subsampled Sketchy Extended (multi-class accuracy).
Our metric learning approach outperforms model regression approaches.
}
\centering
\resizebox{0.95\linewidth}{!}{
\begin{tabular}{lccccc}
\toprule
 & \textbf{w2v} & \multicolumn{2}{c}{\textbf{sketch}} & \multicolumn{2}{c}{\textbf{image}}\\
        &   & \textit{one}-shot & \textit{five}-shot & \textit{one}-shot & \textit{five}-shot\\
\cmidrule(lr){1-1}\cmidrule(lr){2-2}\cmidrule(lr){3-4}\cmidrule(lr){5-6}
M2M~\cite{hu2018sketch} & \deemph{n/a} & \deemph{n/a} & 79.93 & \deemph{n/a} & 93.55 \\
F2M~\cite{hu2018sketch} & 35.90 & 68.16 & 83.01 & 84.12 & 93.89 \\
\rowcolor{Gray}
\textit{This paper} & \textbf{80.39} & \textbf{82.19} & \textbf{85.13} & \textbf{90.63} & \textbf{94.63} \\
\bottomrule
\end{tabular}
}
\label{tab:sota2}
\vspace{-1em}
\end{table}

\subsection{Few-shot sketch-based image classification}\label{sec:sota2}
\paragraph{Setup} Few-shot sketch-based image classification focuses on classifying natural images from one or a few labeled sketches. The few-shot categories have not been observed during training. Different from the zero-shot retrieval scenario, the few-shot classification evaluation has access to the labels of the unseen classes in the evaluation phase. For example, this comes through the form of sketches or word embeddings.
We report results on the \emph{Sketchy Extended} dataset~\cite{sketchy2016,liu2017deep}. 
For fair comparison with Hu~\etal\cite{hu2018sketch}, we subsample the Sketchy Extended to match the size of their private split. We select the same 115 classes for training and 10 classes for testing. We also rely on VGG19~\cite{simonyan2014very} as a backbone. We evaluate the performance with the multi-class accuracy.
Classification is done by measuring the distance to the class prototypes. 
Following Hu~\etal\cite{hu2018sketch}, we evaluate on three different modes by setting the prototypes of the unseen classes to: (i) word vectors (w2v), (ii) \textit{one} or \textit{five} sketch representations, and (iii) \textit{one} or \textit{five} image representations. The latter is considered as an upper-bound of this cross-domain task. Following Hu~\etal~\cite{hu2018sketch}, the model is trained once and we report the average classification accuracy over 500 runs with different sets of sketches or images in the few-shot evaluation.

\paragraph{Results} Table~\ref{tab:sota2} compares our formulation to two baselines introduced by Hu~\etal~\cite{hu2018sketch}.
M2M regresses weights for natural image classification from the weights of the sketch classifier while F2M regresses weights from sketch representations.
For the first evaluation mode, we obtain an accuracy of 76.73\%, compared to 35.90\%, which reiterates the importance of a semantic alignment for categorical cross-domain search. In the few-shot evaluation, the biggest relative improvement is achieved in the one-shot evaluation.
It is also interesting to compare the w2v and one-shot sketch evaluation modes. As the one-shot sketch exhibits a higher score, it means that sketch representations capture visual details that cannot be described with word representations only.
Our approach is also effective for cross-domain classification, especially with low shots.

\paragraph{Qualitative analysis} To understand how to best employ our approach for few-shot sketch-based image classification, we provide the most and least effective sketches for image classification in Figure~\ref{fig:sota2-q}. Since categories are condensed to a single prototypical sketch, our approach desires sketches with details and in canonical configurations. Results are degraded when such assertions are not met. For example, Figure~\ref{fig:few:a} shows a well sketched \say{cat} in one of the canonical positions while Figure~\ref{fig:few:b} exhibits a \say{cat} without any whiskers and in a strange view as we only see the face. Another important assertions is the sketch separability. For example, the \say{airplane} sketch in Figure~\ref{fig:few:b} could be confused with a \say{knife}. Appearance, viewpoint and separability matter when relying on sketches for few-shot image classification.

\begin{figure*}[t]
\vspace{-1em}
\centering
\hfill
\begin{overpic}[width=0.45\linewidth]{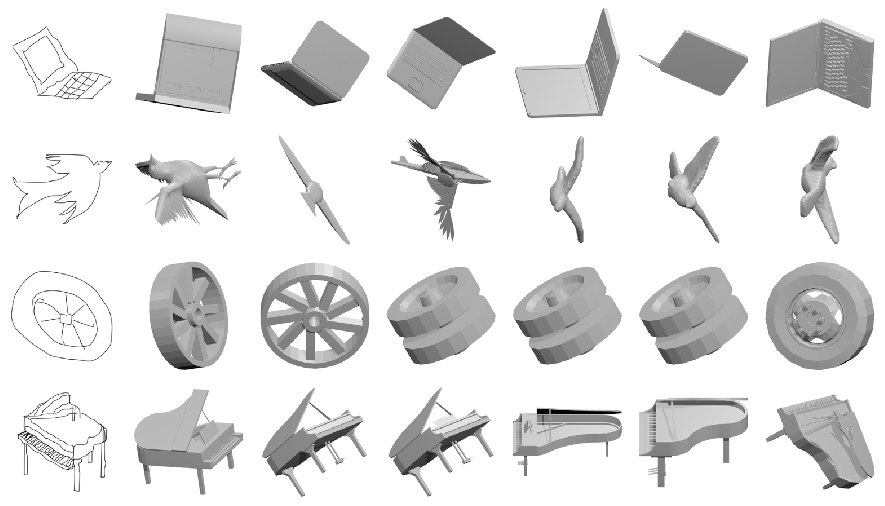}
\put(-2,47){\scriptsize \rotatebox{90}{\textit{laptop}}}
\put(-2,30){\scriptsize \rotatebox{90}{\textit{flying bird}}}
\put(-2,18){\scriptsize \rotatebox{90}{\textit{wheel}}}
\put(-2,4){\scriptsize \rotatebox{90}{\textit{piano}}}
\end{overpic}\hfill
\begin{overpic}[width=0.45\linewidth]{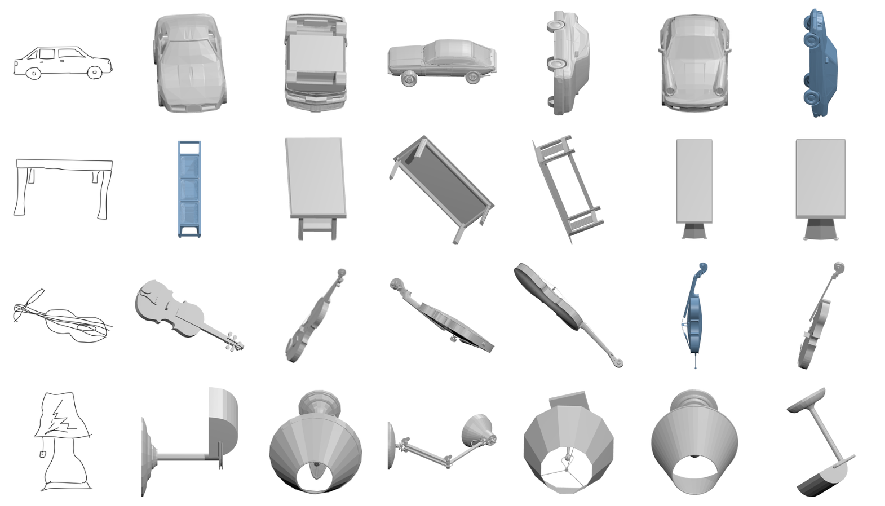}
\put(-2,47){\scriptsize \rotatebox{90}{\textit{sedan}}}
\put(-2,33){\scriptsize \rotatebox{90}{\textit{table}}}
\put(-2,19){\scriptsize \rotatebox{90}{\textit{violin}}}
\put(-2,2){\scriptsize \rotatebox{90}{\textit{tablelamp}}}
\end{overpic}\hfill\null
\caption{ \textbf{Qualitative analysis} of many-shot sketch-based 3D shape retrieval on Part-SHREC14. Incorrect results are shown in \textit{{\color{NavyBlue}blue}}.
Our approach handles the unaligned shapes by projecting all views to the same semantic prototype in the shared space.
An open problem remains the confusion with categories that are close both in semantics and in appearance (\eg, ``violin'' \vs~``cello'').
}
\label{fig:sota3-q}
\vspace{-1em}
\end{figure*}
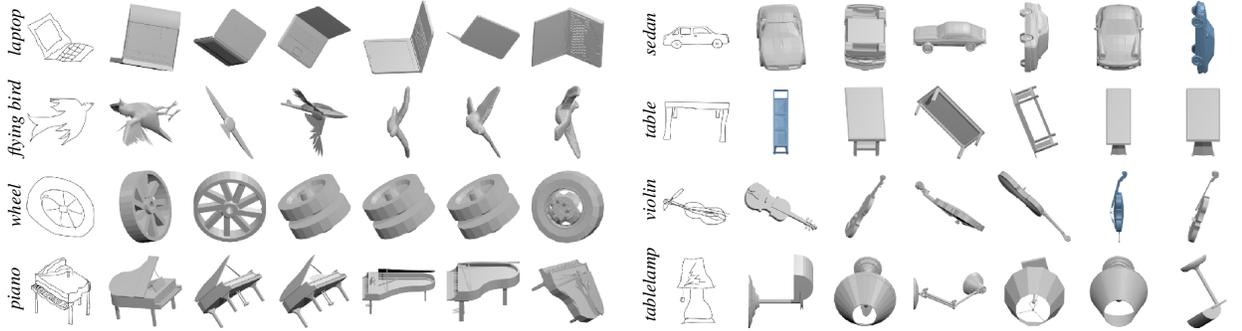

\subsection{Many-shot sketch-based 3D shape retrieval}\label{sec:sota3}
\paragraph{Setup} Sketch-based 3D shape retrieval focuses on retrieving 3D shape models from a sketch query, where both training and testing samples share the same set of classes. We evaluate on three datasets. \emph{SHREC13}~\cite{li2013shrec} is constructed from the TU-Berlin~\cite{eitz2012hdhso} and Princeton Shape Benchmark~\cite{Shilane2004TPS} datasets, resulting in 7,200 sketches and 1,258 3D shapes from 90 classes. The training set contains 50 sketches per class, the testing set 30. \emph{SHREC14}~\cite{li2014shrec} contains more 3D shapes and more classes, resulting in 13,680 sketches and 8,987 3D shapes from 171 classes. The training and testing splits of sketches follow the same protocol as SHREC13. We also report on \emph{Part-SHREC14}~\cite{Qi2018SemanticEF}, which contains 3,840 sketches and 7,238 3D shapes from 48 classes. The sketch splits also follow the same protocol, while the 3D shapes are now split into 5,812 for training and 1,426 for testing to avoid overlap.

\begin{table}[t]

\caption{\textbf{Comparison 3} to many-shot sketch-based 3D shape retrieval on SHREC13, SHREC14, and Part-SHREC14.
Having a metric space revolving around semantic prototypes benefits five out of six metrics.
}
\centering
\vspace{-0.5em}
\subfloat[ {\small SHREC13}\label{tab:shape:sota1}]{
\resizebox{0.95\linewidth}{!}{
\begin{tabular}{lcccccc}
\toprule
 & NN & FT & ST & E & DCG & mAP \\
\cmidrule(lr){1-1}\cmidrule(lr){2-7}
Siamese~\cite{wang2015sketch} & 
0.405 & 0.403 & 0.548 & 0.287 & 0.607 & 0.469
\\
Shape2Vec~\cite{Tasse2016shape} &
0.620 & 0.628 & 0.684 & 0.354 & 0.741 & 0.650
\\
DCML~\cite{dai2017deep} &
0.650 & 0.634 & 0.719 & 0.348 & 0.766 & 0.674
\\
LWBR~\cite{xie2017learning} &
0.712 & 0.725 & 0.785 & 0.369 & 0.814 & 0.752
\\
DCA~\cite{chen2018deep} &
0.783 & 0.796 & 0.829 & 0.376 & 0.856 & 0.813
\\
SEM~\cite{Qi2018SemanticEF} &
0.823 & 0.828 & 0.860 & 0.403 & 0.884 & 0.843
\\
DSSH~\cite{chen2019deep} &
\textbf{0.831} & 0.844 & 0.886 & 0.411 & 0.893 & 0.858
\\
\rowcolor{Gray}
\emph{This paper} &
0.825 & \textbf{0.848} & \textbf{0.899} & \textbf{0.472} & \textbf{0.907} & \textbf{0.865}\\
\bottomrule
\end{tabular}}}
\\\vspace{-5px}
\subfloat[ {\small SHREC14}\label{tab:shape:sota2}]{
\resizebox{0.95\linewidth}{!}{
\begin{tabular}{lcccccc}
\toprule
 & NN & FT & ST & E & DCG & mAP\\
\cmidrule(lr){1-1}\cmidrule(lr){2-7}
Siamese~\cite{wang2015sketch} & 
0.239 & 0.212 & 0.316 & 0.140 & 0.496 & 0.228
\\
Shape2Vec~\cite{Tasse2016shape} &
0.714 & 0.697 & 0.748 & 0.360 & 0.811 & 0.720
\\
DCML~\cite{dai2017deep} &
0.272 & 0.275 & 0.345 & 0.171 & 0.498 & 0.286
\\
LWBR~\cite{xie2017learning} &
0.403 & 0.378 & 0.455 & 0.236 & 0.581 & 0.401
\\
DCA~\cite{chen2018deep} &
0.770 & 0.789 & 0.823 & 0.398 & 0.859 & 0.803
\\
SEM~\cite{Qi2018SemanticEF} &
\textbf{0.804} & 0.749 & 0.813 & 0.395 & 0.870 & 0.780
\\
DSSH~\cite{chen2019deep} &
0.796 & 0.813 & 0.851 & 0.412 & 0.881 & 0.826
\\
\rowcolor{Gray}
\emph{This paper} &
0.789 & \textbf{0.814} & \textbf{0.854} & \textbf{0.561} & \textbf{0.886} & \textbf{0.830}\\
\bottomrule
\end{tabular}}}
\\
\vspace{-5px}
\subfloat[{\small Part-SHREC14}\label{tab:shape:sota3}]{
\resizebox{0.95\linewidth}{!}{
\begin{tabular}{lcccccc}
\toprule
 & NN & FT & ST & E & DCG & mAP\\
\cmidrule(lr){1-1}\cmidrule(lr){2-7}
Siamese~\cite{wang2015sketch} & 0.118 & 0.076 & 0.132 & 0.073 & 0.400 & 0.067 \\
SEM~\cite{Qi2018SemanticEF} & \textbf{0.840} & 0.634 & 0.745 & 0.526 & 0.848 & 0.676 \\
DSSH~\cite{chen2019deep} & 0.838 & 0.777 & 0.848 & 0.624 & 0.888 & 0.806 \\
\rowcolor{Gray}
\emph{This paper} & 0.816 & \textbf{0.799} & \textbf{0.891} & \textbf{0.685} & \textbf{0.910} & \textbf{0.831} \\
\bottomrule
\end{tabular}}}

\label{tab:sota3}
\vspace{-1em}
\end{table}

Following previous works~\cite{chen2018deep,xie2017learning,su15mvcnn}, we generate 2D projections for all 3D shape models using the Phong reflection model~\cite{phong1975illumination}. Similarly, we render 12 different views by placing a virtual camera evenly spaced around the unaligned 3D shape model with an elevation of 30 degrees. We only aggregate the multiple views during testing to reduce complexity.
We report six retrieval metrics~\cite{li2014comparison}. The nearest neighbour (NN) denotes precision@$1$. The first tier (FT) is the recall@$K$, where $K$ is the number of 3D shape models in the gallery set of the same class as the query. The second tier (ST) is the recall@2$K$. The E-measure (E) is the harmonic mean between the precision@32 and the recall@32. The discounted cumulated gain (DCG) and mAP are also reported.

\paragraph{Results} Table~\ref{tab:sota3} shows the results on all three benchmarks and six metrics.
We compare to seven state-of-the-art baselines, which mostly focus on learning a joint feature space of sketches and 3D shapes with metric learning~\cite{hadsell2006dimensionality,chopra2005learning,schroff2015facenet}.
Across all three benchmarks, we observe the same trend, where we obtain the highest scores for five out of the six baselines. Only for the precision@1 metric (NN) do the recent approaches of Chen~\etal\cite{chen2019deep} and Qi~\etal\cite{Qi2018SemanticEF} obtain higher scores on all three benchmarks. 
A first reason for this behaviour is that both approaches directly optimize for the nearest neighbour metric. Qi~\etal\cite{Qi2018SemanticEF} search in the label space while Chen~\etal\cite{chen2019deep} perform a learned hashing. A second reason comes from their usage of more complex 3D shape representations. Qi~\etal\cite{Qi2018SemanticEF} work with point clouds while Chen~\etal\cite{chen2019deep} sample 2D views from various viewpoints.
Our approach, while simple in nature, provides competitive results compared to the current state-of-the-art in many-shot sketch-based 3D shape retrieval.

\paragraph{Qualitative analysis} To gain insight in our approach for retrieving 3D shapes from sketches, we provide qualitative examples in Figure~\ref{fig:sota3-q}. Rotations of unaligned shapes can be handled. For example, 3D shapes of \say{laptop} or \say{piano} are retrieved despite the large differences in rotation angles.
Yet, confusion remains with visually similar categories. This happens when the search needs to differentiate among fine-grained categories. For example, differences are subtle between \say{sedan cars} and \say{sports cars}, or between \say{violin} and \say{cello}.
Although errors can appear with semantically similar categories, our method can retrieve highly variable 3D shapes from sketches.

\section{Conclusion}
In this paper, we open visual search beyond two domains to scale to any number of domains.
This translates into a search between any pair of source and target domains, a search from a combination of multiple sources, or a search within a combination of multiple targets.
This creates new challenges as all domains should map to the same embedding space, while new domains should be able to be incorporated efficiently.
To achieve open cross-domain visual search, we propose a simple approach based on domain-specific prototype learners to align the semantics of multiple visual domains in a common space.
Learning a mapping to a common space enables a visual search among any number of source or target domains. The addition of new domains consists in the training of a new prototype learner, without the need to retrain previous models.
Empirical demonstrations on novel \textit{open} cross-domain visual search tasks present how to search across multiple domains.
State-of-the-art results on existing \textit{closed} cross-domain visual search tasks show the effectiveness of our approach.

{\small{
\paragraph{Acknowledgements}
We thank Herke van Hoof for initial insight, Qing Liu for helpful correspondence, as well as Zenglin Shi and Hubert Banville for feedback. William Thong is partially supported by an NSERC scholarship.}}

{\small
\bibliographystyle{ieee_fullname}
\bibliography{egbib}
}

\end{document}